%% file: main.tex
\documentclass{article}
\usepackage[utf8]{inputenc}
\usepackage{makecell}
\usepackage{colortbl}
\usepackage{multirow}
\usepackage{hhline}
\usepackage{rotating}
\usepackage{xfp}
\usepackage{subcaption}
\usepackage{etoolbox}
\usepackage{wrapfig}
\usepackage{mdframed}
\usepackage{xstring}
\usepackage{url}
\usepackage{xinttools}
\usepackage{afterpage}
\usepackage{tikz}
\usetikzlibrary{shapes.geometric, arrows,decorations.pathreplacing,calc,tikzmark}

\newif\ifen
\newif\ifde

\newcommand{\en}[1]{\ifen#1\fi}
\newcommand{\de}[1]{\ifde#1\fi}

\entrue 

\ifde
  \usepackage[ngerman]{babel}
  \usepackage[german]{authblk}
  
  \usepackage[autostyle,german=quotes]{csquotes}
  \MakeOuterQuote{"}
\fi

\ifen
  \usepackage[]{authblk}
  \usepackage[autostyle, english = american]{csquotes}
  \MakeOuterQuote{"}
\fi

\usepackage[colorlinks]{hyperref}
\hypersetup{linkcolor=black,urlcolor=red,citecolor=olive}

\newcommand{\projectname}{Fairness Compass}

\date{}

\title{
\en{Towards the Right Kind of Fairness in AI}
\de{Welche Art von Fairness macht KI-Systeme gerecht?}\\
\vspace{14pt}
\large 
\en{A guide on the different metrics. And the tool ``\projectname'' to choose the best option for your project.}
\de{Eine Übersicht der verschiedenen Optionen. Und das Tool ,,\projectname'' für die optimale Auswahl.}\\\vspace{14pt}
}

\author{Boris Ruf}
\author{Marcin Detyniecki\thanks{\{boris.ruf,marcin.detyniecki\}@axa.com}}
\affil{AXA GO, AI Research, Paris, France}

\input{macros/confusion_matrices}
\input{macros/predictions}

\newboolean{NR}
\newboolean{BR}
\newboolean{FDR}
\newboolean{FOR}
\newboolean{TPR}
\newboolean{TNR}

\begin{document}

\maketitle

\begin{abstract}
\en{
Fairness is a concept of justice. Various definitions exist, some of them conflicting with each other. In the absence of an uniformly accepted notion of fairness, choosing the right kind for a specific situation has always been a central issue in human history. When it comes to implementing sustainable fairness in artificial intelligence systems, this old question plays a key role once again: How to identify the most appropriate fairness metric for a particular application? The answer is often a matter of context, and the best choice depends on ethical standards and legal requirements. Since ethics guidelines on this topic are kept rather general for now, we aim to provide more hands-on guidance with this document. Therefore, we first structure the complex landscape of existing fairness metrics and explain the different options by example. Furthermore, we propose the "\projectname", a tool which formalises the selection process and makes identifying the most appropriate fairness definition for a given system a simple, straightforward procedure. Because this process also allows to document the reasoning behind the respective decisions, we argue that this approach can help to build trust from the user through explaining and justifying the implemented fairness.}
\de{Fairness ist ein theoretisches Konzept von Gerechtigkeit, für das es verschiedene Definitionen gibt. Manche von ihnen stehen in Konflikt miteinander, und es existiert keine universell gültige Version. Wenn man sicherstellen will, dass Systeme, die auf Künstlicher Intelligenz (KI) beruhen, faire Entscheidungen treffen, ist daher zunächst die Wahl der passenden Definition ausschlaggebend. Bei der idealen Art von Fairness für ein KI-System kommt es in der Regel auf den Kontext an, und die richtige Auswahl wird von ethischen Grundsätzen und gesetzlichen Vorgaben bestimmt. Da die Ethikrichtlinien für diesen Bereich momentan noch sehr allgemein gehalten sind, ist dieser Bericht als praktischer Leitfaden gedacht. Zu diesem Zweck strukturieren wir zunächst das vielschichtige Angebot unterschiedlicher Fairnessdefinitionen. Wir stellen die verschiedenen Optionen vor und erklären sie anhand konkreter Beispiele. Im zweiten Teil des Berichts präsentieren wir den "\projectname" – ein Tool, das den Auswahlprozess formalisiert und aus der Suche nach der optimalen Fairnessdefinition für eine konkrete Anwendung einen klar definierten, transparenten Vorgang macht. Da auf diese Weise auch die Begründungen für die einzelnen Entscheidungen dokumentiert werden können, eignet sich dieses Verfahren auch, um allen beteiligten Parteien die implementierte Art von Fairness zu erklären, und so das Vertrauen in die Anwendung zu stärken.}
\end{abstract}

\newpage
\tableofcontents

\newpage

\section{\en{Introduction}\de{Einführung}}
\en{The last few years have seen a number of tremendous breakthroughs in the field of artificial intelligence (AI). A significant part of this success is due to major advances in machine learning (ML), the data analytics technique behind AI. Machine learning does recognise correlations in large data sets. Due to its ability to process loads of information in short time, it can uncover statistical patterns in data that humans cannot spot. This gives access to new kind of insights from the data which allow improved data analysis and model predictions. Although not absolutely error-free, the results clearly outperform conventional approaches, and often even human experts. The areas of application are extensive and include medical diagnosis, university admission, loan allocation, recidivism prediction, recruitment, online advertisement, face recognition, language translation, recommendation engines, fraud detection, credit limits, pricing and false news detection. 

The heavy dependence on data poses a new challenge though. The data used for training a machine learning algorithm are considered the \emph{ground truth}. This means that during the learning phase, these data constitute the comprehensive representation of the real world which the algorithm seeks to approximate. If the training data includes any kind of unwanted bias, the resulting algorithm will incorporate and enforce it. Worse still, in the absence of robust explanations for the results, it is hardly possible for humans to recognise biased predictions of machine learning algorithms as such.

Unwanted bias may happen to be directed against sensitive subgroups, defined for instance by gender, ethnicity or age. As a consequence, people from one such group would be generally disadvantaged by the system. However, systematic unequal treatment of individuals from different sensitive groups is considered discrimination, and there is broad consensus in our society that making a distinction based on a personal characteristic which is usually not a matter of choice is unfair. Hence, anti-discrimination laws in plenty of legislations prohibit actions of this nature.

The traditional approach to fight discrimination in statistical models when using deterministic algorithms is known as "anti-classification". This principle is firmly encoded in current legal standards and it simply rules to exclude any attribute which defines membership in a sensitive subgroup as feature from the data. For example, a user's gender may not be collected and processed in many scenarios. However, since machine learning is backed by "big data" which contain highly correlated features that can serve as possible proxies for those sensitive attributes, this approach has been shown to be insufficient to avoid discrimination in AI systems~\cite{Corbett2018}.

Two main sources for undesired bias have been identified in the machine learning pipeline. First, if the training data are incorrect or not sufficiently representative in certain aspects, this fault may become the source of correlations which do not exist in this form in reality. In such a case, the machine learning algorithm may detect patterns which are in fact not meaningful. Second, the training data may indeed faithfully represent the real world, but the status quo does not appear ideal. Without correction, the machine learning algorithm would reproduce the current state and thus manifest an existing shortcoming. The objective is therefore to adjust for this bias in the resulting algorithm.

Whatever the source, plenty of mitigation techniques have been presented by researchers lately to deal with bias in data and make AI applications more fair. This is an encouraging development towards maintaining trust in AI and eventually overcoming some of the potentially biased human judgments which impair automatic decision-making. Besides the technical task of adjusting the algorithms or the data, an equally important philosophical question needs to be settled: what kind of fairness is the objective? Fairness is a concept of justice and a variety of definitions exist which sometimes conflict with each other. Hence, there is no uniformly accepted notion of fairness available. In fact, the most appropriate fairness definition depends on the use case and it is often a matter of legal requirements and ethical standards.

The purpose of this document is to assist AI stakeholders in settling for the desired ethical principles by questions and examples. Applying such a procedure will not only help to identify the best fairness definition for a given AI application, but it will also make the choice transparent and the implemented fairness more understandable for the end user.

In the remainder, we first introduce some mathematical basics which are useful to assess and compare the performance of machine learning algorithms. Second, we explain the problem of unwanted bias in data. Next, we present the most commonly used fairness definitions in research and explain the ethical principles they stand for. Afterwards, we illustrate by example how these fairness definitions may be mutually contradictory. Finally, we present the "\projectname" which constitutes an actionable guide for AI stakeholders to translate ethical principles into fairness definitions.}

\de{Im Laufe der letzten Jahre konnten bedeutende Durchbrüche im Bereich der Künstlichen Intelligenz (KI) erzielt werden. Ein großer Anteil an diesen Erfolgen ist auf Fortschritte beim Machinellen Lernen (ML) zurückzuführen, der Schlüsseltechnologie hinter kognitiven Systemen. Machinelles Lernen beschreibt die Fähigkeit einer Maschine, in großen Datensätzen Korrelationen zu erkennen. Aufgrund seiner Eigenschaft, riesige Mengen an Informationen in kurzer Zeit zu analysieren, können statistische Muster in Daten erkannt werden, die dem menschlichen Auge verborgen bleiben. Diese wiederum ermöglichen neuartige Erkenntnisse aus den Daten, welche die Datenanalyse und Modellprognosen optimieren helfen. Obwohl auch diese Ergebnisse nicht komplett fehlerfrei sind, so übertreffen sie doch die herkömmlichen Ansätze, und schneiden oft sogar besser ab als menschliche Experten. Die Anwendungsgebiete für diese Technik sind vielfältig und umfassen medizinische Diagnosen, Studienplatzvergaben, Kreditbewilligungen, Rückfallprognosen, Rekrutierung, Onlinewerbung, Gesichtserkennung, Sprachübersetzung, Empfehlungssysteme, Betrugserkennung, Kreditkartenlimits, Preisgestaltung und Faktencheck in Sozialen Netzwerken.

Die große Abhängigkeit von den Daten birgt allerdings eine neue Herausforderung. Die Daten, die zum Training eines Algorithmus herangezogen werden, betrachtet man als \emph{Ground Truth}. Das bedeutet, dass diese Daten während der Lernphase die vollumfängliche Realität abbilden, die das Prognosemodell anzunähern sucht. Sollten die Trainingsdaten auf irgendeine Weise unerwünschte Verzerrungen (\textit{bias}) aufweisen, wird der trainierte Algorithmus diese widerspiegeln und sogar verstärken. Und weil die Logik von KI-Algorithmen für Menschen bislang nicht verständlich erklärbar ist, lassen sich diese Verzerrungen weder im Modell noch im Ergebnis ohne Weiteres als solche erkennen.

Von unerwünschten Verzerrungen können Teile der Gesellschaft betroffen sein, die sich über sensible Merkmale wie zum Beispiel das Geschlecht, die ethnische Herkunft oder das Alter definieren. Demzufolge kann es vorkommen, dass Menschen aus diesen Gruppen von einem KI-System kategorisch benachteiligt werden. Systematische, ungleiche Behandlung von Individuen mit verschiedenen sensiblen Merkmalen wird als Diskriminierung betrachtet, und es herrscht breiter Konsens in unserer Gesellschaft, dass es unfair ist, wenn Ungleichbehandlung auf Grundlage persönlicher Eigenschaften stattfindet, auf welche der oder die Betroffene in der Regel keinen Einfluss hat. Entsprechend verbieten Antidiskriminierungsgesetze in vielen Ländern derartiges Handeln.

Bei statistischen Modellen, die mit traditionellen, deterministischen Algorithmen erzeugt werden, wird zur Vermeidung von Diskriminierung ein Verfahren eingesetzt, das als "Anti-Classification" bekannt ist. Dieses Prinzip ist auch in der aktuellen Gesetzgebung verankert und sieht einfach vor, dass Datenparameter, die sensible Merkmale beschreiben, von der Verwendung ausgeschlossen sind. So darf zum Beispiel das Geschlecht einer Person in vielen Anwendungsfällen weder erhoben noch verwendet werden. Nun basiert Maschinelles Lernen allerdings auf "Big Data", welche äußerst komplexe Korrelationen enthalten. Diese haben zur Folge, dass mitunter Parameter, welche unbedenklich scheinen und nicht als "sensibel" eingestuft sind, indirekt sensible Informationen preisgeben können. Auf Grundlage dieser Verflechtungen konnte demonstriert werden, dass für KI-Systeme ein solcher Ansatz zur Bekämpfung von Diskriminierung ungenügend ist~\cite{Corbett2018}.

Im Entwicklungsprozess von ML-Modellen wurden zwei Hauptquellen für unerwünschte Verzerrungen festgestellt. Erstens kann es vorkommen, dass die Trainingsdaten fehlerhaft oder in bestimmter Hinsicht unzureichend repräsentativ sind. Derartige Mängel können die Ursache für Korrelationen in den Daten sein, die auf diese Weise in der Realität nicht zu finden sind. In so einem Fall erkennt der Algorithmus ein Muster, welches eigentlich keine Bedeutung hat. Zweitens ist es möglich, dass die Trainingsdaten zwar durchaus die Wirklichkeit abbilden, dieser Status Quo aber nicht der idealen Zielvorstellung entspricht. Wenn keine Korrektur erfolgt, reproduziert der Algorithmus den aktuellen Zustand und verfestigt so den bestehenden Missstand. Das Ziel besteht unter diesen Umständen darin, dass der finale Algorithmus die vorhandenen Verzerrungen ausgleicht.

In den letzten Jahren haben Forscherinnen und Forscher zahlreiche Methoden entwickelt, die diese Verzerrungen in den Daten, egal welchen Ursprungs, abschwächen, und KI-Systeme so fairer machen können. Das ist eine ermutigende Entwicklung, die hoffentlich das Vertrauen in KI stärkt und an deren Ende manche potentiell voreingenommene, menschliche Entscheidungen möglicherweise von unparteiischen, automatischen Entscheidungen ersetzt werden können. Neben der technischen Herausforderung, die Algorithmen oder die Daten anzupassen, gilt es allerdings eine ebenso wichtige, philosophische Frage zu klären: Welche Art von Fairness ist das Ziel? Fairness ist ein theoretisches Konzept von Gerechtigkeit, und es existieren verschiedene Definitionen, von denen manche untereinander in Konflikt stehen. Es gibt also keine universell anwendbare Form von Fairness, die allen Vorstellungen gleichermaßen genügt. Die optimale Definition hängt vielmehr vom konkreten Anwendungsfall ab und wird meistens von ethischen Grundsätzen und gesetzlichen Rahmenbedingungen bestimmt.

Dieses Dokument soll die Verantwortlichen für KI-Systemen unterstützen, die angestrebten ethischen Prinzipien anhand von Fragen und Beispielen festzulegen. Das vorgeschlagene Verfahren vereinfacht dabei nicht nur die Wahl der besten Fairnessdefinition für eine bestimmte Anwendung, sondern es macht diese Auswahl auch transparent und die implementierte Fairness für alle Beteiligten besser nachvollziehbar. 

Die nachfolgenden Kapitel sind wie folgt strukturiert. Zunächst führen wir einige mathematische Grundlagen ein, die nützlich sind, um die Eigenschaften von Algorithmen des Maschinellen Lernens zu bewerten und zu vergleichen. Dann vertiefen wir das Problem von Verzerrungen in Daten. Anschließend präsentieren wir die am häufigsten verwendeten Definitionen von Fairness in der Forschung und erläutern die ethischen Prinzipien, die sie repräsentieren. Im folgenden Kapitel illustrieren wir beispielhaft, wie sich diese Fairnessdefinitionen mitunter gegenseitig widersprechen. Schließlich führen wir den "\projectname" ein: unser praxisnahes Tool für KI-Entscheider, mit dem sich angestrebte ethische Standards in die passende Fairnessdefinition übersetzen lassen.}

\section{\en{Fundamentals}\de{Grundlagen}}
\en{For better understanding of the following sections, we introduce here some fundamental knowledge about machine learning, and a few statistical measures commonly used to characterise its performance.}

\de{Um die nachstehenden Kapitel besser verstehen zu können, vermitteln wir hier zunächst gewisse Grundkenntnisse zu Maschinellem Lernen und führen außerdem einige statistische Maße ein, die zur Prüfung und Bewertung der Systeme nützlich sind. }

\subsection{\en{Machine Learning}\de{Maschinelles Lernen}}
\begin{figure}[h]
  \centering
  \includegraphics[width=0.5\linewidth]{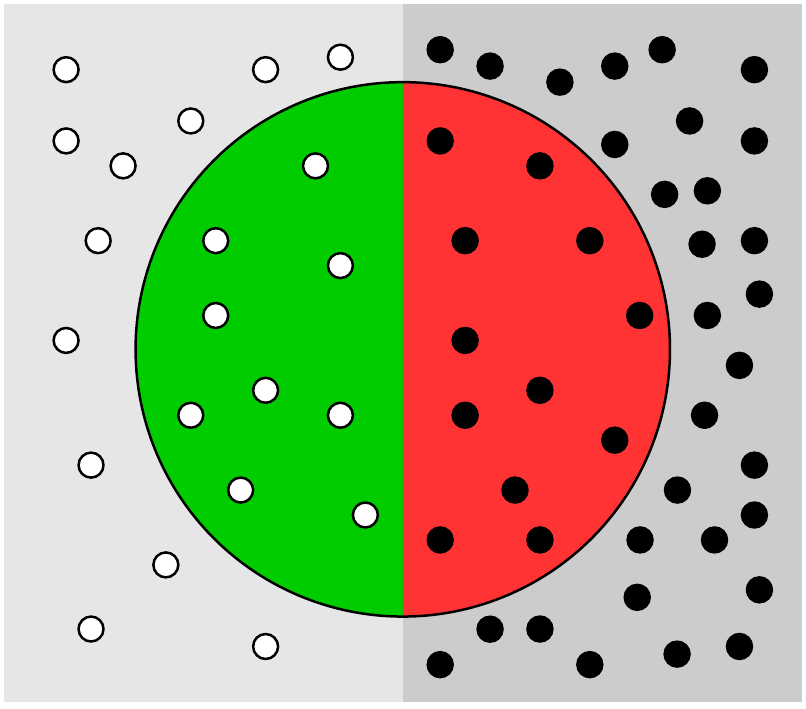}
  \caption{\en{Graphical representation of the sample scenario. White dots represent positive samples, black ones negative samples. The big circle constitutes the classifier.}\de{Grafische Darstellung der Ergebnisse aus \autoref{table:predictions}. Weiße Punkte repräsentieren unrechtmäßige Fälle (1), schwarze Punkte stehen für legitime Fälle (0). Der große Kreis stellt das Vorhersagemodell dar.}}
  \label{fig:graphical_representation}
\end{figure}

\en{Compared with traditional programming, the difference of machine learning is that the reasoning behind the algorithm’s decision-making is not defined by hard-coded rules which were explicitly programmed by a human, but it is rather learned by example data: Thousands, sometimes millions of parameters get optimised without human intervention to finally capture a generalised pattern of the data. The resulting model allows to make predictions on new, unseen data with high accuracy.

This approach can be used for two different kinds of problems: On the one hand for classification, where the task is to predict discrete classes such as categories, for example. On the other hand for regression, where the objective is to predict a continuous quantity, for instance a price. Throughout this document, we only consider classification tasks, and for the sake of simplicity, we focus on the binary case with two classes: positive (1) or negative (0). For model output we either consider those very class labels 0 and 1, or a score $S$ which corresponds to the probability for the sample to be positive.

To illustrate the concepts in this document, we introduce a sample scenario about fraud detection in insurance claims. This fictional setting will serve as a running example throughout the following sections. Verifying the legitimacy of an insurance claim is essential to prevent abuse. However, fraud investigations are labour intensive for the insurance company. In addition, for some types of insurance, many claims may occur at the same time – for example, due to natural disasters that affect entire regions. For policyholders, on the other hand, supplementary checks can be annoying, for example when they are asked to answer further questions or provide additional documents. Both parties are interested in a quick decision: The customers expect timely remedy, and the company tries to keep the effort low. Therefore, an AI system that speeds up such a task could prove very useful. Concretely, it should be able to reliably identify legitimate insurance claims in order to make prompt payment possible. Potentially fraudulent cases should also be reliably detected and flagged for further investigation.

In order to analyse the performance of a classifier, we compare the predicted output $\widehat{Y}$ with the true output value $Y$. In the claims data, the output value 1 stands for a fraudulent claim, while 0 represents a legitimate claim. \autoref{table:predictions} shows sample predictions for our running example. For better illustration, we also provide a graphical representation of the same results in \autoref{fig:graphical_representation}. The white dots correspond to the positive samples ($Y=1$), here actual fraudulent claims. The black dots represent negative samples ($Y=0$), actual legitimate claims in the present scenario. The big circle constitutes the boundary of the classifier: Dots within the circle have been predicted as positive/fraudulent ($\widehat{Y}=1$), dots outside the circle as negative/legitimate ($\widehat{Y}=0$). The different background colours further show where the classifier was right (green and dark grey), and where not (red and light grey).}

\de{Maschinelles Lernen unterscheidet sich von traditioneller Programmierung dadurch, dass die Logik des Algorithmus nicht auf hartkodierten Regeln beruht, die von einem Menschen so explizit festgelegt wurden, sondern vielmehr anhand von Beispielen erlernt wird: Tausenden, manchmal sogar Millionen Parameter werden ohne menschliches Eingreifen optimiert, um letztlich ein strukturelles Muster aus den Daten abzubilden. Das resultierende Prognosemodell ist dann im Stande, für neue Datensätze aus demselben Anwendungsbereich Vorhersagen mit hoher Präzision zu treffen.

Dieser Ansatz kann für zwei unterschiedliche Arten von Problemen eingesetzt werden: Für Klassifikation, wo die Aufgabe darin besteht, diskrete Klassen vorherzusagen, wie etwa Kategorien. Und für Regression, wo das Ziel ist, einen kontinuierlichen Wert zu prognostizieren, wie zum Beispiel einen Preis. Im vorliegenden Bericht gehen wir ausschließlich auf Klassifikationsprobleme ein, und der Einfachheit halber konzentrieren wir uns auf den binären Fall mit zwei Klassen: positiv (1) oder negativ (0). Als Ausgabewerte des Modells erwarten wir entweder eben jene Label 0 und 1; oder einen Score $S$, welcher die Wahrscheinlichkeit ausdrückt, dass eine Instanz positiv ist.

Um die hier vorgestellten Konzepte besser veranschaulichen zu können, verwenden wir ein fiktionales Szenario aus dem Bereich der Betrugserkennung bei Versicherungsfällen. Im Laufe der folgenden Kapitel werden wir uns immer wieder auf dieses Anwendungsbeispiel beziehen. Die Prüfung der Legitimität eines Versicherungsfalls ist unerlässlich um Missbrauch zu verhindern. Allerdings handelt es sich dabei für das Versicherungsunternehmen um einen aufwendigen und personalintensiven Vorgang. Zudem treten für manche Versicherungsarten viele Schadensfälle zeitgleich ein – etwa durch Naturkatastrophen, die ganze Regionen betreffen. Für Versicherungsnehmer wiederum können genaue Kontrollen lästig sein, zum Beispiel wenn sie gebeten werden, weitere Fragen zu beantworten oder zusätzliche Dokumente nachzureichen. Beiden Parteien ist dabei an einer raschen Entscheidung gelegen: Die Kunden erwarten schnelle Abhilfe, und das Unternehmen versucht den Aufwand gering zu halten. Ein KI-System, das eine solche Aufgabe beschleunigt, könnte sich also als sehr nützlich erweisen. Konkret sollte es im Stande sein, rechtmäßige Versicherungsfälle sicher zu erkennen, um eine zeitnahe Auszahlung möglich zu machen. Potentiell betrügerische Fälle sollten ebenfalls zuverlässig entdeckt und für weitere Ermittlungen gekennzeichnet werden. 

Zur Bewertung der Qualität eines Vorhersagemodells vergleichen wir die Prognosen $\widehat{Y}$ mit den wahren Klassen $Y$. Im System unseres Anwendungsszenarios ist dabei ein legitimer Versicherungsfall mit dem Wert 0 kodiert, und ein Betrugsversuch mit 1. \autoref{table:predictions} enhält einige Beispielvorhersagen, die wir in \autoref{fig:graphical_representation} grafisch umsetzen, um die Zahlen besser verständlich zu machen: Die weißen Punkte entsprechen den positiven Datensätzen ($Y=1$), im Beispielkontext die tatsächlich betrügerischen Versicherungsfälle. Die schwarzen Punkte repräsentieren die negativen Instanzen ($Y=0$), hier also die legitimen Fälle. Der große Kreis stellt die Grenzen des Vorhersagemodells dar: Punkte innerhalb des Kreises wurden als positiv/betrügerisch klassifiziert ($\widehat{Y}=1$), Punkte außerhalb als negativ/legitim ($\widehat{Y}=0$). Die unterschiedlichen Hintergrundfarben machen sichtbar, wo das Modell richtig lag (grün und dunkelgrau), und wo nicht (rot und hellgrau).}

\def\a{9}
\def\b{12}
\def\c{12}
\def\d{30}

\begin{table}[h]
  \predictions{\a}{\b}{\c}{\d}
  \caption{\en{Tabular evaluation of the sample scenario: true output class $Y$ compared to predictions $\widehat{Y}$}\de{Beispieldaten für das Versicherungsszenario. Legitime Fälle sind mit 0 kodiert, unrechtmäßige mit 1. In der oberen Zeile stehen die wahren Klassen $Y$, in der unteren die zugehörigen Prognosen $\widehat{Y}$ des Modells.}}
  \label{table:predictions}
\end{table}

\en{It is worth noting that in this oversimplified 2-dimensional example, drawing an ideal border which separates black and white dots and thus defining a perfect classifier would be obvious. In high-dimensional real world use cases, however, it is hardly possible to obtain a perfect classifier with error rates of zero; optimisation always remains a matter of trade-offs.}

\de{Selbstverständlich wäre es in diesem stark vereinfachten zweidimensionalen Beispiel nicht weiter schwer, die perfekte Grenze zwischen schwarzen und weißen Punkten zu ziehen und damit ein ideales Vorhersagemodell zu erhalten. In echten Applikationen, die oft tausende Aspekte in Form von Datenparametern berücksichtigen, ist es allerdings kaum möglich, ein perfektes, fehlerfreies Modell zu finden; Kompromisse sind bei der Modellbildung unumgänglich.}

\subsection{\en{Statistical Measures}\de{Statistische Kennzahlen}}\label{ssec:statistical_measures}
\en{A so-called "confusion matrix" helps to visualise and compute statistical measures commonly used to inspect the performance of a machine learning model. The rows of the matrix represent the actual output classes, in our case 0 or 1. The columns represent the predicted output classes by the given classifier. The cells where the predicted class corresponds to the actual class contain the counts of the correctly classified instances. Wherever the classes differ, the classifier got it wrong and the numbers represent incorrectly classified samples.}

\de{Die sogenannte "Wahrheitsmatrix" ist ein nützliches Mittel, um statistische Kennzahlen, die häufig zur Bewertung von Vorhersagemodellen herangezogen werden, darzustellen und zu berechnen. Die Zeilen der Matrix repräsentieren dabei die wahren Klassen, in unserem Fall 0 oder 1. Die Spalten beziehen sich auf die Vorhersagen des Modells. In den Zellen, wo die vorhergesagte Klasse mit dem tatsächlichen Ausgabewert übereinstimmt, stehen die Summen der jeweils korrekt klassifizierten Datensätze. Wo sich die Klassen unterscheiden, lag das Modell in seiner Vorhersage falsch und die Zellen enthalten die Summen der entsprechend fehlerhaft eingeordneten Fälle.}

\begin{table}[h]
  \confusionmatrixterms
  \caption{\en{Schema of confusion matrix}\de{Schema einer Wahrheitsmatrix}}
  \label{table:confusionmatrix_schema}
\end{table}

\en{On an abstract level, the figures in the cells are generally identified by the terms provided in \autoref{table:confusionmatrix_schema}. Taking the data from our running example in \autoref{table:predictions} as a basis, the related confusion matrix looks like \autoref{table:confusionmatrix_example}. We notice that the given classifier correctly predicted \a\ claims to be fraudulent and \d\ claims to be legitimate. However, it also falsely predicted \b\ claims to be legitimate, which were in fact fraudulent, and \c\ claims to be fraudulent, which really were not.}

\de{Auf abstrakter Ebene werden die Werte aus den Zellen üblicherweise mit den Begriffen aus \autoref{table:confusionmatrix_schema} bezeichnet. Wenn wir die Daten aus unserem laufenden Beispiel in \autoref{table:predictions} als Grundlage nehmen, ergibt sich die Wahrheitsmatrix in \autoref{table:confusionmatrix_example}. Wir stellen fest, dass das Vorhersagemodell in diesem Beispiel \a\ Versicherungsfälle korrekt als betrügerisch einstuft, und \d\ ebenfalls richtig als legitim. Es klassifiziert jedoch auch \b\ weitere Fälle als legitim, die eigentlich widerrechtlich sind, und \c\ rechtmäßige Fälle zu Unrecht als Betrugsversuch.
}

\begin{table}[ht]
  \confusionmatrixshort{\a}{\b}{\c}{\d}
  \caption{\en{Confusion matrix for the sample data from \autoref{table:predictions}}\de{Resultierende Wahrheitsmatrix aus den Beispieldaten in \autoref{table:predictions}}}
  \label{table:confusionmatrix_example}%
\end{table}

\en{Revisiting the illustration in \autoref{fig:graphical_representation}, we further realise that the coloured segments in the schema correspond to the different cells in the confusion matrix: false negatives (light grey), true positives (green), false positives (red), and true negatives (dark grey).

From the confusion matrix we can extract plenty of interesting statistical measures. We describe those measures in the text and provide their formulas and graphical representations in \autoref{table:metrics}.}

\de{Wenn wir uns erneut die grafische Darstellung in  \autoref{fig:graphical_representation} vor Augen führen sehen wir nun, dass die farbig hinterlegten Segmente den unterschiedlichen Zellen in der Wahrheitsmatrix entsprechen: Falsch-negative Prognosen (hellgrau), richtig-positive Prognosen (grün), falsch-positive Prognosen (rot) und richtig-negative Prognosen (dunkelgrau).

Der Wahrheitsmatrix lassen sich zahlreiche statistische Kennzahlen entnehmen, die für eine Analyse des Modells interessant sind. Wir beschreiben diese im Einzelnen im nachfolgenden Text und stellen in \autoref{table:metrics} außerdem ihre Formeln und grafischen Darstellungen bereit.}

\begin{table}[]
    \centering
    \begin{tabular}{ | m{13em} | m{10em}| m{3em} | } 
        \hline
        \en{Actual positives}\de{Tatsächlich positive Fälle} & $P=FN+TP$ & 
        \begin{center}
        \hspace*{0.05in}
        \includegraphics[height=0.05\textwidth]{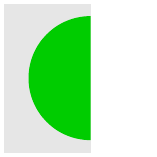}
        \vspace*{-0.1in}
        \end{center}\\
        \hline
         \en{Actual negatives}\de{Tatsächlich negative Fälle} & $N=FP+TN$ & 
        \begin{center}
        \hspace*{-0.2in}
        \vspace*{-0.1in}
        \includegraphics[height=0.05\textwidth]{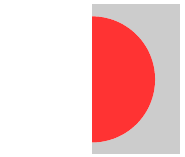}
        \end{center}\\
        \hline
         \en{Base rate}\de{Basisrate} & $BR=\frac{P}{P+N}$ & 
         \vspace*{-0.11in}
         \begin{center}
         \includegraphics[height=0.09\textwidth]{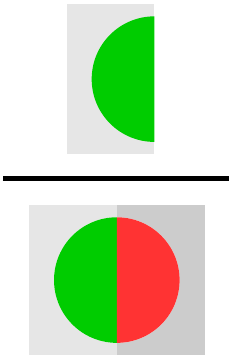} 
         \vspace*{-0.18in}
         \end{center}\\
         \hline
         \en{Positive rate}\de{Positivrate} & $PR=\frac{TP+FP}{P+N}$ &  \vspace*{-0.11in}
         \begin{center}
         \includegraphics[height=0.09\textwidth]{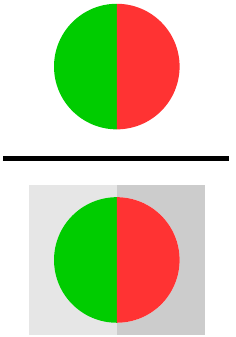} 
         \vspace*{-0.18in}
         \end{center}\\
         \hline
         \en{Negative rate}\de{Negativrate} & $NR=\frac{TN+FN}{P+N}$ &  \vspace*{-0.11in}
         \begin{center}
         \includegraphics[height=0.09\textwidth]{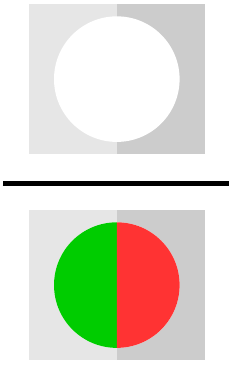} 
         \vspace*{-0.18in}
         \end{center}\\
         \hline
         \en{Accuracy}\de{Korrektklassifikationsrate (\textit{Accuracy})} & $ACC=\frac{TP+TN}{P+N}$ &  \vspace*{-0.11in}
         \begin{center}
         \includegraphics[height=0.09\textwidth]{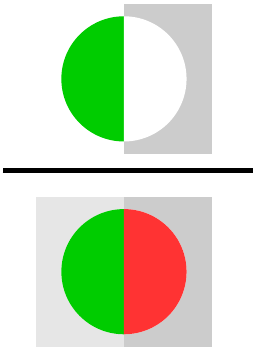} 
         \vspace*{-0.18in}
         \end{center}\\
         \hline
         \en{Misclassification rate}\de{Fehlklassifikationsrate (\textit{Misclassification rate})}& $MR=\frac{FN+FP}{P+N}$ &  \vspace*{-0.11in}
         \begin{center}
         \includegraphics[height=0.09\textwidth]{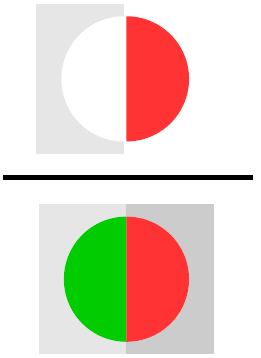} 
         \vspace*{-0.18in}
         \end{center}\\
         \hline
         \en{True positive rate}\de{Richtig-positiv-Rate\newline(\textit{True positive rate})} & $TPR=\frac{TP}{P}$ &  \vspace*{-0.11in}
         \begin{center}
         \includegraphics[height=0.09\textwidth]{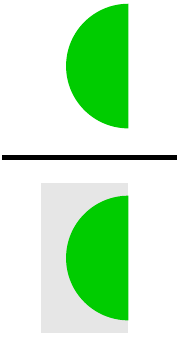} 
         \vspace*{-0.18in}
         \end{center}\\
         \hline
         \en{True negative rate}\de{Richtig-negativ-Rate\newline(\textit{True negative rate})} & $TNR=\frac{TN}{N}$ &  \vspace*{-0.11in}
         \begin{center}
         \includegraphics[height=0.09\textwidth]{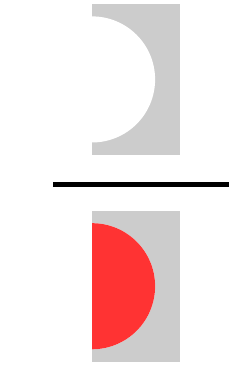} 
         \vspace*{-0.18in}
         \end{center}\\
         \hline
         \en{False positive rate}\de{Falsch-positiv-Rate} & $FPR=\frac{FP}{N}$ &  \vspace*{-0.11in}
         \begin{center}
         \includegraphics[height=0.09\textwidth]{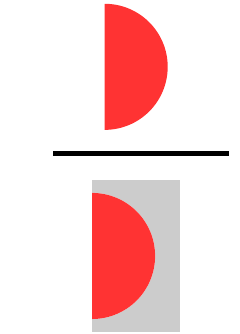} 
         \vspace*{-0.18in}
         \end{center}\\
         \hline
         \en{False negative rate}\de{Falsch-negativ-Rate} & $FNR=\frac{FN}{P}$ &  \vspace*{-0.11in}
         \begin{center}
         \includegraphics[height=0.09\textwidth]{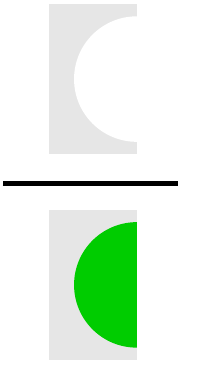} 
         \vspace*{-0.18in}
         \end{center}\\
         \hline
         \en{False discovery rate}\de{Falscherkennungsrate\newline(\textit{False discovery rate})} & $FDR=\frac{FP}{TP+FP}$ &  \vspace*{-0.11in}
         \begin{center}
         \includegraphics[height=0.09\textwidth]{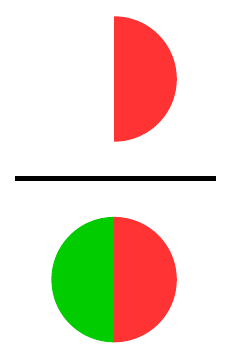} 
         \vspace*{-0.18in}
         \end{center}\\
         \hline
         \en{Positive predictive value}\de{Positiver Vorhersagewert\newline(\textit{Positive predictive value})} & $PPV=\frac{TP}{TP+FP}$ &  \vspace*{-0.11in}
         \begin{center}
         \includegraphics[height=0.09\textwidth]{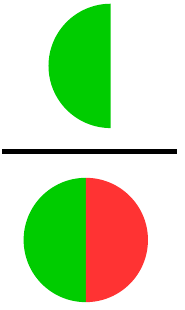} 
         \vspace*{-0.18in}
         \end{center}\\
         \hline
         \en{False omission rate}\de{Falschauslassungsrate\newline(\textit{False omission rate})} & $FOR=\frac{FN}{TN+FN}$ &  \vspace*{-0.11in}
         \begin{center}
         \includegraphics[height=0.09\textwidth]{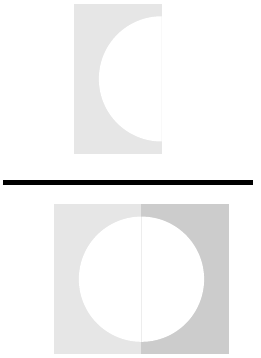} 
         \vspace*{-0.18in}
         \end{center}\\
         \hline
         \en{Negative predictive value}\de{Negativer Vorhersagewert\newline(\textit{Negative predictive value})} & $NPV=\frac{TN}{TN+FN}$ &  \vspace*{-0.11in}
         \begin{center}
         \includegraphics[height=0.09\textwidth]{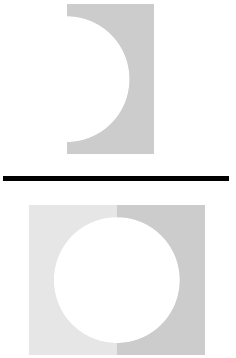} 
         \vspace*{-0.18in}
         \end{center}\\
         \hline
    \end{tabular}
    \caption{\en{Metrics derived from confusion matrix}\de{Abgeleitete Formeln aus der Wahrheitsmatrix}}
    \label{table:metrics}
\end{table}

\en{First, we count the \textbf{actual positives} in the data set. This number is the sum of the true positives and the false negatives, which can be viewed as missed true positives. Likewise, the number of \textbf{actual negatives} is the sum of the true negatives and the false positives, which again can be viewed as missed true negatives. In our example, those figures represent the numbers of actual fraudulent claims and actual legitimate claims.

The (positive) \textbf{base rate}, sometimes also called the prevalence rate, represents the proportion of actual positives with respect to the entire data set. In our example, this rate describes the share of actual fraudulent claims in the data set.

The \textbf{positive rate} is the overall rate of positively classified instances, including both correct and incorrect decisions. The \textbf{negative rate} is the ratio of negative classification, again irrespective of whether the decisions were correct or incorrect. In our example, the positive rate is the rate of all claims suspected to be fraudulent, and the negative rate represents the rate of claims which are predicted to be legitimate.

\textbf{Accuracy} is the ratio of the correctly classified instances (positive and negative) of all decisions. In return, the \textbf{misclassification rate} is the ratio of the misclassified instances over all decisions. In our example, accuracy is the proportion of claims which were correctly classified, either as fraudulent or as legitimate. The misclassification rate refers to the failed classifications, the proportion of incorrect decisions taken by the classifier.

The \textbf{true positive rate} and the \textbf{true negative rate} describe the proportions of correctly classified positive and negative instances, respectively, of their actual occurrences. In the example, the true positive rate describes the share of all actual fraudulent claims which were detected as such. The true negative rate is the share of actual legitimate claims which were successfully discovered.

Directly linked, the \textbf{false positive rate} and the \textbf{false negative rate} describe the error rates. The false positive rate denotes the proportion of actual negatives which was falsely classified as positive. In the same way, the false negative rate describes the proportion of actual positives which was misclassified as negative. In our example, the false positive rate is the proportion of all actual legitimate claims which were falsely classified as fraudulent. On the other way around, the false negative rate is the proportion of all actual fraudulent claims which slipped through the system and were falsely classified as legitimate.

The \textbf{false discovery rate} describes the share of misclassified positive classifications of all positive predictions. So, it is about the proportion of positively classified instances which were falsely identified or discovered as such. On the contrary, the \textbf{false omission rate} describes the proportion of false negative predictions of all negative predictions. These instances, which are actually positive, were overlooked – they were mistakenly passed over or omitted. In our example, the false discovery rate is the error rate of all claims which were classified as fraudulent. The false omission rate describes the share of actually fraudulent claims of all claims which were classified as legitimate. 

In a similar approach, but rather focusing on the correctly classified samples, the \textbf{positive} and the \textbf{negative predictive values} describe the ratio of samples which were correctly classified as positive or negative from all the positive or negative predictions. In the example, the positive predictive value is the proportion of correctly identified claims in all claims which were flagged as fraudulent. The negative predictive value is the proportion of correctly classified claims in all claims which were flagged as legitimate.} 

\de{Zunächst ermitteln wir die \textbf{tatsächlich positiven Fälle} im Datensatz. Diese Zahl ergibt sich aus den Summen der richtig-positiven und der falsch-negativen Prognosen. Letztere können auch als verfehlte positive Prognosen betrachtet werden. Entsprechend ergibt sich die Zahl der \textbf{tatsächlich negativen Fälle} aus der Summe der richtig-negativen und der falsch-positiven Prognosen, welche wiederum als verpasste negative Prognosen verstanden werden können. In unserem Beispiel entsprechen diese Kennzahlen den jeweiligen Summen der tatsächlich betrügerischen und der tatsächlich legitimen Versicherungsfälle.

Die (positive) \textbf{Basisrate}, manchmal auch als Prävalenzrate bezeichnet, steht für den Anteil der tatsächlich positiven Fälle, bezogen auf den kompletten Datensatz. Im Beispiel beschreibt diese Rate den wahren Anteil betrügerischer Fälle im Datensatz.

Die \textbf{Positivrate} wiederum ist der proportionale Anteil der positiven Bescheide, unabhängig davon, ob die Entscheidung richtig oder falsch war. Die \textbf{Negativrate} beschreibt umgekehrt die Rate der Negativebescheide, wieder ungeachtet dessen, ob die Entscheidung korrekt oder inkorrekt war. In unserem Beispiel entspricht die Positivrate der Rate jener Fälle, die als betrügerisch eingestuft wurden. Die Negativrate bezeichnet den Anteil der als legitim klassifizierten Fälle.

Die \textbf{Korrektklassifikationsrate} (\textit{accuracy}) ist die Erfolgsquote, die den Anteil der korrekten Prognosen (positiv und negativ) von allen Entscheidungen bemisst. Im Gegenzug definiert die \textbf{Fehlklassifikationsrate} (\textit{misclassification rate}) den Anteil der Fehlentscheidungen. In unserem Anwendungsbeispiel gibt die Korrektklassifikationsrate den Anteil der zurecht als legitim und der zurecht als betrügerisch eingeordneten Fälle wieder. Die Fehlklassifikationsrate bezieht sich auf die Fehlentscheidungen – der Anteil jener Instanzen also, bei der sich das Modell geirrt hat.

Die \textbf{Richtig-positiv-Rate} und die \textbf{Richtig-negativ-Rate} beschreiben die Proportionen der korrekt positiv bzw. korrekt negativ eingeordneten Instanzen, anteilig ihrer tatsächlichen Vorkommnisse. Im Beispiel steht die Richtig-positiv-Rate für den Anteil der tatsächlich betrügerischen Fälle, der vom Modell als solche erkannte wurde. Die Richtig-negativ-Rate beschreibt die Rate der tatsächlich legitimen Fälle, die erfolgreich als solche eingeordnet wurden.

In direktem Zusammenhang dazu beschreiben die \textbf{Falsch-positiv-Rate} und die \textbf{Falsch-negativ-Rate} die Fehlerquoten. Die Falsch-positiv-Rate bezeichnet den Anteil der eigentlich negativen Instanzen, der fälschlicherweise positiv klassifiziert wurde. Gleichermaßen steht die Falsch-negativ-Rate für die Rate der eigentlich positiven Fälle, die fälschlicherweise als negativ eingestuft wurden. Im Beispielszenario steht die Falsch-positiv-Rate für den Anteil der legitimen Fälle, welcher irrtümlich als betrügerisch klassifiziert wurde. Umgekehrt ist die Falsch-negativ-Rate der Anteil von den eigentlich betrügerischen Fällen, der vom System "übersehen" und inkorrekt als legitim klassifiziert wurde.

Die \textbf{Falscherkennungsrate} (\textit{false discovery rate}) bezeichnet den Anteil der Fehlentscheidungen von allen positiv beschiedenen Fällen. Es geht also um den Anteil jener Instanzen, die zu Unrecht als positiv identifiziert bzw. entdeckt wurden. Andersherum beschreibt die \textbf{Falschauslassungsrate} (\textit{false omission rate}) die Rate der fälschlicherweise negativ eingeordneten Fälle von allen Negativbescheiden. Diese eigentlich positiven Instanzen wurden ignoriert, d.h. sie wurden irrtümlich übergangen bzw. ausgelassen. Im Beispiel entspricht die Falscherkennungsrate der Fehlerrate von den Fällen, die als betrügerisch eingestuft wurden. Die Falschauslassungsrate bezeichnet hingegen die Fehlerrate von den als legitim vorhergesagten Versicherungsfälle – wie viele Fälle also irregulär als rechtmäßig eingestuft wurden. 

Ähnlich, aber mit Schwerpunkt auf die korrekt klassifizierten Instanzen, beschreiben der \textbf{positive} und der \textbf{negative Vorhersagewert} den jeweiligen Anteil der positiven bzw. negativen Vorhersagen, der richtig war. Im Beispiel ist der positive Vorhersagewert der Anteil von jenen Prognosen, die einen Betrug vermuten, und damit richtig liegen. Der negative Vorhersagewert wiederum ist der Anteil der Fälle, die als legitim prognostiziert wurden, und das zurecht.} 

\section{\en{Problem of Bias}\de{Das Problem von Verzerrungen}}
\en{Up until here, we have analysed the data as one population and did not consider the possible existence of sensitive subgroups in the data. However, since decisions from machine learning algorithms often affect humans, many data sets contain sensitive subgroups by nature of the data. Such subgroups may for example be defined by gender, race or religion. The membership of an instance is usually identified by a sensitive attribute $A$. To analyse potential bias of a classifier, we split the results by this sensitive attribute into subgroups and investigate possible discrepancies among them. Any such deviation could be an indicator for discrimination against one sensitive group.

The idea of pursuing fairness on the basis of membership in one or several sensitive groups is called "group fairness"~\cite{mehrabi2019}. This approach is also adopted in anti-discrimination laws in many legislations with varying lists of sensitive attributes~\cite{EU2012,Barocas2016}. In addition, another concept exists in research which tries to achieve "individual fairness" by aiming at similar treatment of similar individuals, taking any attribute into account~\cite{Dwork2011}. In the scope of this document, we focus on group fairness, and to facilitate matters, we only consider two different sensitive subgroups. Therefore, we assume one sensitive binary attribute $A$ which can take the values 0 or 1, for instance representing the gender.

Unwanted bias is said to occur when the statistical measures described in the previous section significantly differ from one sensitive subgroup to another. Without any closer analysis on a per-subgroup basis, such a problem can go completely unnoticed. Please note that in order to "see" the groups in the data the sensitive information is obviously required to be available.

We now examine our running example on insurance fraud detection for unwanted bias. The output from the trained model remains unchanged, but this time we assume two sensitive subgroups in the data, specified by the sensitive attribute $A$. For instance, we split the data into men ($A$=0) and women ($A$=1). The separate confusion matrices for each subgroup in \autoref{table:confusionmatrices_separated} enable us to compare the performance measures.}

\de{Bis jetzt haben wir für die statistische Analyse immer die Daten als Ganzes zugrunde gelegt, und nicht weiter berücksichtigt, dass der Datensatz aus sensiblen Untergruppen bestehen könnte. Die Entscheidungen von ML-Algorithmen betreffen jedoch oft Menschen. Daher ist es schon aufgrund der Beschaffenheit der Daten naheliegend, dass diese unterschiedliche demographische Gruppen umfassen, beispielsweise definiert durch das Geschlecht einer Person, deren ethnischem Hintergrund oder Konfession. Technisch wird die Zugehörigkeit zu einer solchen Gruppe meistens durch ein sensibles Attribut $A$ im Datensatz festgehalten. Um ein Prognosemodell auf mögliche Verzerrungen zu prüfen, unterteilen wir die Ergebnisse mithilfe dieses sensiblen Attributs in verschiedene Datensätze und untersuchen deren statistische Merkmale auf Abweichungen. Unterschiedliche Kennwerte können Anzeichen von strukturellen Fehlern (\textit{biases}) sein – verzerrte Prognosen also, die eine Ungleichbehandlung der sensiblen Untergruppen bedeutet. 

Die Idee, Fairness auf Grundlage der Zugehörigkeit zu einer oder mehreren sensiblen Gruppen anzustreben, wird "Gruppenfairness" genannt~\cite{mehrabi2019}. Dieser Ansatz findet sich auch in den Antidiskriminierungsgesetzen zahlreicher Gesetzgebungen wieder, mit verschiedenen Listen geschützter, sensibler Attribute~\cite{EU2012,Barocas2016}. Alternativ gibt es in der Forschung ein weiteres Konzept names "Individuelle Fairness", das stattdessen eine Gleichbehandlung von Individuen angestrebt, die sich in sämtlichen Attributen ähneln – sensibel und nicht-sensibel~\cite{Dwork2011}. Im Rahmen dieses Dokuments konzentrieren wir uns auf das Konzept der Gruppenfairness, und der Einfachheit halber gehen wir außerdem nur von zwei sensiblen Untergruppen aus. Folglich enthält unser Datensatz nur ein binäres, sensibles Attribute $A$, das die Werte 0 oder 1 annehmen kann, zum Beispiel um das Geschlecht zu kodieren.

Von unerwünschten Verzerrungen ist die Rede, wenn einige der statistischen Kennzahlen aus dem vorherigen Kapitel zwischen den Gruppen wesentlich abweichen. Es ist also notwendig, auf Basis der sensiblen Untergruppen separate Datenanalysen durchzuführen, was im Übrigen die Verfügbarkeit der sensiblen Attribute voraussetzt. Anderenfalls lassen sich Probleme dieser Art nur schwer erkennen.

Wir untersuchen nun unser Anwendungsbeispiel zum Thema Betrugserkennung bei Versicherungsfällen auf Verzerrungen. Die Prognosedaten des trainierten Modells bleiben dabei unverändert, allerdings betrachten wir die Daten jetzt für zwei sensible Untergruppen, die durch das sensible Attribute $A$ definiert sind. Zu diesem Zweck unterteilen wir die Daten für Männer ($A$=0) und Frauen ($A$=1). Anhand der unterschiedlichen Wahrheitsmatrizen für diese Untergruppen in \autoref{table:confusionmatrices_separated} können wir jetzt deren statistische Eigenschaften inspizieren.}

\begin{table}[ht]
  \begin{subfigure}{.48\textwidth}
    \confusionmatrix{7}{7}{6}{22}{$A$=0}
    \caption{\en{Men}\de{Männer}}
  \end{subfigure}
  \hfill
  \begin{subfigure}{.48\textwidth}
    \confusionmatrix{2}{5}{6}{8}{$A$=1}
    \caption{\en{Women}\de{Frauen}}
  \end{subfigure}
  \caption{\en{Separate confusion matrices for sensitive subgroups}\de{Unterschiedliche Wahrheitsmatrizen für sensible Untergruppen}}
  \label{table:confusionmatrices_separated}
\end{table}

\en{We notice that the base rates (BR) are identical in both subgroups which means in this example that men and women are equally likely to file a fraudulent (or a legitimate) claim. However, the true negative rate (TNR) for men is 0.79, while for women it is 0.57. This means 79\% of the valid claims filed by men get correctly classified as legitimate, while for women that’s the case for only 57\% of the same type of claims. On the other hand, the false omission rate for men is 24\% and for women it is 38\%. So, fraudulent claims filed by women have a higher chance to remain undetected than fraudulent claims filed by men.}

\de{Wir stellen fest, dass die Basisraten (BR) für beide Gruppen identisch sind, was bedeutet, dass die Wahrscheinlichkeit, dass Männer und Frauen einen betrügerischen (oder einen legitimen) Schadensfall melden, gleich hoch ist. Die Richtig-negativ-Rate (TNR) liegt für Männer allerdings bei 0.79, während sie für Frauen 0.57 beträgt. Das bedeutet, dass 79\% aller legitimen Fälle, die von Männern eingereicht wurden, korrekt als legitim eingeordnet wurden, wohingegen für Frauen das nur für 57\% der Fälle des gleichen Typs gilt. Andererseits liegt die Falschauslassungsrate (FOR) für Männer bei 24\% und für Frauen bei 38\%. Betrügerische Fälle, die von Frauen eingereicht werden, haben also eine höhere Chance unerkannt zu bleiben, als betrügerische Fälle von Männern.}

\section{\en{Available Fairness Definitions}\de{Verfügbare Fairnessdefinitionen}}
\en{The problem of biased AI has attracted attention only recently, but the research community has already produced several fairness definitions to measure unwanted bias in outputs of machine learning models as described in the previous section. Additionally, plenty of mitigation methods have been proposed to ensure the kind of fairness they represent. For more details on the different mitigation approaches we refer the interested reader to survey papers on the subject as a starting point~\cite{Corbett2018,mehrabi2019}. In this document, we focus on the definitions of fairness and their impact on the results in real world scenarios. 

In the following, we present the most commonly used definitions for group fairness and explain their characteristics by example. Since all notions relate to one of three fundamental conditions of statistical independence which are commonly known as independence, sufficiency, and separation, we segment the definitions by those categories~\cite{barocas2019}.}

\de{Obwohl das Problem von Verzerrungen in KI-Systemen erst seit wenigen Jahren diskutiert wird, hat die Forschungsgemeinschaft bereits zahlreiche Lösungs\-vorschläge präsentiert. Dazu zählen etliche Fairnessdefinitionen, mit denen sich unerwünschte Verzerrungen in den Ergebnissen von Prognosemodellen, wie wir sie im vorherigen Kapitel beschrieben haben, messen lassen. Außerdem wurden verschiedene Methoden präsentiert, mit denen sich die jeweilige Art von Fairness erwirken lässt. Für weiterführende Informationen zu diesen Methoden verweisen wir auf entsprechende Übersichtsartikel als Ausgangspunkt~\cite{Corbett2018,mehrabi2019}. In diesem Bericht konzentrieren wir uns auf die Fairnessdefinitionen und deren Auswirkungen auf die Ergebnisse in realen Anwendungsszenarien. 

Im weiteren Verlauf stellen wir die am häufigsten verwendeten Fairnessdefinitionen für Gruppenfairness vor, und erklären deren Eigenschaften an Beispielen. Alle Definitionen lassen sich einem von drei statistischen Prinzipien zuordnen, die wir im Folgenden als Überkategorien verwenden: Die "bedingungslose" Unabhängigkeit, sowie die bedingten Unabhängigkeiten Suffizienz und Separierung~\cite{barocas2019}.}

\subsection{\en{Independence}\de{Unabhängigkeit}}\label{ssec:independence}
\en{Statistically, fairness definitions satisfy independence if the sensitive attribute $A$ is unconditionally independent of the prediction $\widehat{Y}$. Practically, this means that when considering all predictions made, the share of positive and negative decisions is proportionally equal among the two sensitive subgroups. On an individual level, this means that the likelihood of being classified as one of the classes is equal for two individuals with different sensitive attributes.}

\de{Statistisch betrachtet erfüllen Fairnessdefinitionen das Prinzip der Unabhängigkeit, wenn das sensible Attribut $A$ von der Prognose $\widehat{Y}$ unbedingt unabhängig ist. Praktisch bedeutet das, dass auf alle Prognosen bezogen, der Anteil positiver und negativer Entscheidungen zwischen den sensiblen Gruppen proportional gleich ist. Auf individueller Ebene gilt dann für zwei Personen mit verschiedenen sensiblen Attributen, dass es für sie gleich wahrscheinlich ist, eine der beiden Klassen zugewiesen zu bekommen.}

\subsubsection{Demographic Parity}\label{ssec:demographic_parity}
\en{The goal of demographic parity is that the favourable outcome should be assigned to each subgroup of a sensitive class at equal rates~\cite{Dwork2011}.

In our running sample scenario, this objective translates to equal rates of negative predictions (=classifications as legitimate) for any claims submitted by men or women. In statistical terms, the negative rates (NR) of both subgroups should be identical. However, for the distributions above, NR=0.42 for men and NR=0.67 for women. We notice a gap of 25 percent points for the favourable outcome between the two sensitive subgroups.

The confusion matrices in \autoref{table:confusionmatrices_demographicparity} show possible results for a new model which was optimised for demographic parity. The number of negative predictions has increased for men, the distribution for women remains unchanged. Both confusion matrices now feature a NR of 0.67. Therefore, demographic parity was successfully achieved. It is not surprising though that manipulating the distribution for men also changes true positive and true negative rates.}

\de{Das Ziel von Demographic Parity ist es, den sensiblen Untergruppen das vorteilhaftere Ergebnis in gleichen Raten zuzuweisen~\cite{Dwork2011}.

In unserem Beispiel ist das negative Ergebnis (=Klassifikation als legitimer Fall) das vorteilhaftere. Demographic Parity verlangt also negative Entscheidungen zu gleichen Raten für Männer und Frauen. Statistisch betrachtet müssen die Negativraten (NR) beider Untergruppen identisch sein. In der vorliegenden Verteilung (\autoref{table:confusionmatrices_separated}) gilt allerdings für Männer NR=0.42 und für Frauen NR=0.67. Wir stellen also eine Abweichung von 25 Prozentpunkten für das vorteilhaftere Ergebnis zwischen den beiden sensiblen Untergruppen fest.

Die Wahrheitsmatrizen in \autoref{table:confusionmatrices_demographicparity} enthalten eine mögliche Verteilung der Ergebnisse eines neuen Modells, das für Demographic Parity optimiert wurde. Die Zahl der negativen Prognosen für Männer ist gestiegen, die Matrix für Frauen bleibt unverändert. Beide Wahrheitsmatrizen weisen jetzt eine NR von 0.67 aus. Insofern wurde Demographic Parity erzielt. Allerdings ist es wenig überraschend, dass die Manipulation der Verteilung der Männer auch Ände\-rungen der Richtig-positiv-Rate und der Richtig-negativ-Rate zur Folge hat.}

\begin{table}[ht]
  \begin{subfigure}{.48\textwidth}
    \setboolean{NR}{true} 
    \confusionmatrix{3}{9}{9}{15}{$A$=0}
    \caption{\en{Men}\de{Männer}}
  \end{subfigure}
  \hfill
  \begin{subfigure}{.48\textwidth}
    \setboolean{NR}{true} 
    \confusionmatrix{6}{3}{3}{15}{$A$=1}
    \caption{\en{Women}\de{Frauen}}
  \end{subfigure}
  \caption{\en{Optimised for Demographic Parity}\de{Optimiert für Demographic Parity}}
  \label{table:confusionmatrices_demographicparity}
\end{table}

\subsubsection{Conditional Statistical Parity}\label{ssec:conditional_statistical_parity}
\en{This definition extends demographic parity by allowing a set of legitimate factors to affect the prediction. The definition is satisfied if members in both subgroups have equal probabilities of being assigned to the favourable outcome while controlling for a set of legitimate attributes~\cite{Kamiran2013}.

In our example, a person's history of prior convictions for fraud could be a legitimate attribute affecting the probability for a claim to be investigated. In this case, an attribute which documents previous attempts of fraud could serve as explaining variable.}

\de{Diese Definition erweitert Demographic Parity durch eine Reihe vordefinierter Attribute, deren Einfluss auf die Prognose als legitim betrachtet wird. Das Ziel ist bei dieser Fairnessdefinition erreicht, wenn beide Untergruppen dieselben Chancen auf das vorteilhaftere Ergebnis haben, nachdem die Effekte der legitimen Kontrollvariablen herausgerechnet wurden~\cite{Kamiran2013}.

In unserem Beispiel könnten vorherige Betrugsversuche eines Versicherungsnehmers die Wahrscheinlichkeit einer genaueren Untersuchung erhöhen. In diesem Fall wäre ein Attribut, das etwaige frühere Betrugsversuche dokumentiert, eine passende Kontrollvariable.}

\subsubsection{Equal Selection Parity}\label{ssec:equal_selection_parity}
\en{While demographic parity seeks to obtain equal rates of a positive outcome, proportional towards the group size, the objective of equal selection parity is to have equal absolute numbers of favourable outcomes across the groups, independent of their group sizes~\cite{Saleiro2018}.

In the example about fraud detection, this fairness definition would be satisfied if the exact same number of cases were to be identified as legitimate from both groups, even when one group had filed more cases than the other.}

\de{Während Demographic Parity gleiche Raten proportional zu den Gruppengrößen anstrebt, ist es das Ziel von Equal Selection Parity, dass jeder Untergruppe das bevorzugte Ergebnis in absoluten Zahlen gleich oft zugeteilt wird – unabhängig von den Gruppengrößen~\cite{Saleiro2018}.

Im Beispiel der Betrugserkennung wäre diese Fairnessdefinition erfüllt, wenn für Männer und Frauen die gleiche Anzahl von Schadensfällen als legitim akzeptiert würde, selbst wenn eine Gruppe insgesamt mehr Fälle gemeldet hat, als die andere.}

\subsection{\en{Sufficiency}\de{Suffizienz}}\label{ssec:sufficiency}
\en{Fairness notions satisfy the statistical concept of sufficiency when the sensitive attribute $A$ is conditionally independent of the true output value $Y$ given the predicted output $\widehat{Y}$. In other words, when considering all positive and negative predictions, the share of correct decisions is equal for both sensitive subgroups. On the individual level this translates to equal chances for individuals with identical predictions but different sensitive attributes to have obtained the right label.}

\de{Fairnesskonzepte erfüllen das Prinzip der Suffizienz, wenn das sensible Attribut $A$ bedingt unabhängig von der wahren Klasse $Y$ gegeben die Prognose $\widehat{Y}$ ist. Für die Prognosen einer jeden Klasse gilt also jeweils, dass $A$ von $Y$ unabhängig ist. In anderen Worten: Für alle positiven und alle negativen Vorhersagen ist der jeweilige Anteil korrekter Entscheidungen für beide sensiblen Untergruppen gleich. Auf individueller Basis lässt sich feststellen, dass Personen, die dieselbe Prognose erhalten haben, aber aus unterschiedlichen sensiblen Gruppen stammen, mit der gleichen Wahrscheinlichkeit der richtigen Klasse zugewiesen wurden.}

\subsubsection{Conditional Use Accuracy Equality}\label{ssec:conditional_use_accuracy_equality}
\en{This fairness definition conditions on the algorithm's predicted outcome, not on the actual outcome~\cite{Berk2016}. In statistical terms this means that the positive predictive value (PPV) and the negative predictive value (NPV) across both groups should be equal. 

In the context of our example, the objective of this fairness definition is that for the claims which were predicted as fraud, the proportion of correct predictions should be equal across all groups. Likewise, for the claims which were predicted as legitimate, the proportion of correct predictions should be the same.}

\de{Entsprechend des Suffizienzkriteriums richtet sich diese Fairnessdefinition an der Prognose des Modells aus~\cite{Berk2016}. Statistisch betrachtet werden der positive Vorhersagewert (PPV) und der negative Vorhersagewert (NPV) für beide Gruppen angeglichen. 

Im Zusammenhang mit unserem Beispiel heißt das, dass für alle Versicherungsfälle, die als betrügerisch klassifiziert wurden, diese Prognose für beide Untergruppen zu gleichen Teilen korrekt sein sollte. Und für die als legitim eingestuften Fälle sollten diese Entscheidungen entsprechend für beide Gruppen gleichermaßen korrekt sein.}

\subsubsection{Predictive Parity}\label{ssec:predictive_parity}
\en{Predictive parity is a relaxed version of conditional use accuracy equality which only conditions on the positive predicted outcome~\cite{Chouldechova2016}. Hence, this fairness definition is already satisfied when only the positive predictive value (PPV) is equal for both groups.}

\de{Bei Predictive Parity handelt es sich um eine abgeschwächte Form von Conditional Use Accuracy Equality, bei der die bedingte Unabhängigkeit nur den positiven  Erwartungswert einschließt~\cite{Chouldechova2016}. Folglich ist diese Fairnessdefinition bereits erfüllt, wenn bloß der positive Vorhersagewert (PPV) für beide Gruppen gleich ist.}

\subsubsection{Calibration}\label{ssec:calibration}
\en{Calibration is similar to conditional use accuracy equality but instead of the binary output it conditions on the predicted probability score $S$. The objective is again to obtain equal positive and negative predictive values for all sensitive groups~\cite{Crowson2016}. Such a form of calibration across subgroups corresponds to equal probabilities of correct (or incorrect) classification and can therefore be achieved by aligning false discovery and false omission rates. 

In the context of our example, calibrating the predictions of the classifier would result in equal chances for men and women to get their legitimate claims investigated without cause or to have their fraudulent claims falsely approved. 

The two confusion matrices in \autoref{table:confusionmatrices_calibration} show the results of the classifier after calibration. The distribution for men was adjusted to match the one for women, which was not modified. Due to the equal base rates in both distributions, this action has also aligned all other statistical measures.}

\de{Calibration ist vergleichbar mit Conditional Use Accuracy Equality, wobei als bedingter Erwartungswert anstatt der binären Klassen ein Score $S$ verwendet wird, der die Wahrscheinlichkeit einer Zuordnung zur positiven Klasse ausdrückt. Das Ziel ist wieder, für alle Untergruppen gleiche positive Vorhersagewerte (PPV) und gleiche negative Vorhersagewerte (NPV) zu erreichen~\cite{Crowson2016}. Diese Form von Kalibrierung der Ergebnisse für beide Untergruppen kann mit identischen Wahrscheinlichkeiten einer korrekten (oder inkorrekten) Klassifizierung gleichgesetzt und entsprechend auch über eine Angleichung der Falscherkennungsrate (FDR) und der Falschauslassungsrate (FOR) erreicht werden.

Im Rahmen unseres Beispiels hat eine Kalibrierung der Modellvorhersagen zur Folge, dass Männer und Frauen die gleichen Chancen haben, dass ihre eigentlich legitimen Fälle als betrügerisch eingestuft, oder dass unrechtmäßige Fälle fälschlicherweise akzeptiert werden. 

Die beiden Wahrheitsmatrizen in \autoref{table:confusionmatrices_calibration} enthalten die kalibrierten Ergebnisse. Die Verteilung der Männer wurde angepasst, damit die FDR und die FOR den Werten der Frauen entspricht. Die Verteilung der Frauen wurde nicht verändert. Augrund der zugrundeliegenden identischen Basisraten in beiden Verteilungen hat diese Operation auch alle anderen statistischen Kennzahlen angeglichen.}
\begin{table}[ht]
  \begin{subfigure}{.48\textwidth}
    \setboolean{FDR}{true} 
    \setboolean{FOR}{true} 
    \confusionmatrix{8}{4}{4}{20}{$A$=0}
    \caption{\en{Men}\de{Männer}}
  \end{subfigure}
  \hfill
  \begin{subfigure}{.48\textwidth}
    \setboolean{FDR}{true} 
    \setboolean{FOR}{true} 
    \confusionmatrix{6}{3}{3}{15}{$A$=1}
    \caption{\en{Women}\de{Frauen}}
  \end{subfigure}
  \caption{\en{Optimised for Calibration}\de{Optimiert für Calibration}}
  \label{table:confusionmatrices_calibration}
\end{table}

\subsection{\en{Separation}\de{Separierung}}\label{ssec:separation}
\en{Fairness definitions satisfy the principle of separation if the sensitive attribute $A$ is conditionally independent of the predicted output $\widehat{Y}$ given the true output value $Y$. This means that among both classes, the proportions of correct predictions are equal per sensitive subgroup. On an individual basis, this condition ensures that two individuals who actually belong to the same class but have different sensitive attributes share the same chance to obtain a correct prediction.}

\de{Fairnessdefinitionen erfüllen das Prinzip der Separierung, wenn das sensible Attribut $A$ bedingt unabhängig vom Vorhersagewert $\widehat{Y}$ gegeben die wahre Klasse $Y$ ist. Die Unabhängigkeit zwischen $A$ und $\widehat{Y}$ ist also gegeben, wenn man die wahren Klassen separat betrachtet. Für sie gilt, dass die Anteile der korrekten Vorhersagen für beide sensible Untergruppen gleich sind. Individuell betrachtet garantiert diese Eigenschaft, dass zwei Personen, die zur gleichen Klasse gehören aber Mitglieder von verschiedenen sensiblen Untergruppen sind, die gleiche Chance auf eine richtige Einordnung haben.}

\subsubsection{Equalised Odds}\label{ssec:equalised_odds}
\en{Another fairness definition called equalised odds aims at equal true positive and true negative rates~\cite{Hardt2016}. The reasoning behind this concept is that the probabilities of being correctly classified should be the same for everyone.

In our recurring example, pursuing equalised odds means that the chances for claims to be correctly classified as legitimate or fraudulent should be equal for men and women; the classifier should not be more or less accurate for one of the subgroups.

In \autoref{table:confusionmatrices_equalisedodds}, we show a possible outcome for our example after optimising for equalised odds. The results for men have been reshuffled to match the true positive and true negative rates of the women. Since the base rates are equal in both subgroups, all other statistical measures have been aligned, too, by this operation.}

\de{Die Fairnessdefinition Equalised Odds erzielt Separierung: Für die wahren Klassen ist jeweils Unabhängigkeit gewährleistet, indem die Richtig-positiv und Richtig-negativ-Raten für die sensible Untergruppen gleich sind~\cite{Hardt2016}. Die Über\-legung hinter diesem Konzept ist, dass die Chancen auf eine korrekte Klassifikation für alle Personen gleich sein sollten.

Auf unser wiederkehrendes Beispiel bezogen bedeutet Equalised Odds, dass für Männer und Frauen die Chancen gleich sind, dass ihre Versicherungsfälle zurecht als legitim oder betrügerisch eingeordnet werden; das Prognosemodell sollte nicht für eine Untergruppe mehr oder weniger präzise funktionieren, als für die andere.

\autoref{table:confusionmatrices_equalisedodds} zeigt ein mögliches Ergebnis für unsere Beispielverteilungen, das Equalised Odds erfüllt. Die Prognosen für Männer wurden angepasst, damit die Richtig-positiv-Rate (TPR) und die Richtig-negativ-Rate (TNR) denen der Frauen entsprechen. Da beide Gruppen die gleichen Basisraten haben, hat dieser Vorgang zur Folge, dass sich auch die übrigen statistischen Kennzahlen angeglichen haben.}

\begin{table}[ht]
  \begin{subfigure}{.48\textwidth}
    \setboolean{TPR}{true}
    \setboolean{TNR}{true} 
    \confusionmatrix{8}{4}{4}{20}{$A$=0}
    \caption{\en{Men}\de{Männer}}
  \end{subfigure}
  \hfill
  \begin{subfigure}{.48\textwidth}
    \setboolean{TPR}{true} 
    \setboolean{TNR}{true} 
    \confusionmatrix{6}{3}{3}{15}{$A$=1}
    \caption{\en{Women}\de{Frauen}}
  \end{subfigure}
  \caption{\en{Optimised for Equalised Odds}\de{Optimiert für Equalised Odds}}
  \label{table:confusionmatrices_equalisedodds}
\end{table}

\subsubsection{Equalised Opportunities}\label{ssec:equalised_opportunities}
\en{Optimising for equalised odds can be a difficult task with more complex, real data, therefore the fairness definition equalised opportunities was proposed as more practicable alternative~\cite{Hardt2016}. In this relaxed version of equalised odds, only the error rates for the positive outcome are required to be equal.

In our example, equalised opportunities is achieved when men and women who filed fraudulent claims are exposed at same rates. For legitimate claims, the rates may differ.}

\de{Bei komplexen Daten aus echten Anwendungen kann sich das Erzielen von Equalised Odds als schwierig erweisen. Daher wurde die Fairnessdefinition Equalised Opportunities als praktikablere Alternative vorgeschlagen~\cite{Hardt2016}. In dieser abgeschwächten Version von Equalised Odds muss lediglich die Fehlerrate für die positive Prognose identisch sein.

Im Beispielkontext ist Equalised Opportunities erfüllt, wenn tatsächlich betrügerische Fälle von Männern und Frauen zu gleichen Raten abgelehnt werden. Für legitime Fälle darf die Rate der akzeptierten Fälle zwischen den beiden Gruppen abweichen.}

\subsubsection{Predictive Equality}\label{ssec:predictive_equality}
\en{Another relaxation of equalised odds is predictive equality. Here, only the error rates for the negative outcome are required to be equal~\cite{Corbett2017}.

In our example, predictive equality is satisfied when men and women can expect their legitimate claims to get classified as legitimate at equal rates. The error rates for fraudulent claims may differ between the subgroups for this fairness definition.

As before, the confusion matrix for women in \autoref{table:confusionmatrices_predicitveequality} is unchanged. For men, the output was modified in order to harmonise the false positive rate among the subgroups. The error rates for the unfavourable outcome of being suspect of fraud still deviate depending on the gender.}

\de{Eine weitere Abwandlung von Equalised Odds ist Predictive Equality. Hierbei müssen sich nur die Fehlerraten für die negative Prognose entsprechen~\cite{Corbett2017}.

In unserem Beispiel ist Predictive Equality erfüllt, wenn Männer und Frauen erwarten können, dass ihre legitimen Versicherungsfälle zu gleichen Raten genehmigt werden. Die Fehlerraten für betrügerische Fälle können indessen bei dieser Fairnessdefinition für die beiden Untergruppen abweichen.

Wie zuvor bleibt die Wahrheitsmatrix für Frauen in \autoref{table:confusionmatrices_predicitveequality} unverändert. Für Männer wurde die Verteilung dahingehend angepasst, dass die Falsch-positiv-Rate (TNR) mit der der Frauen übereinstimmt. Die Fehlerraten für das unvorteilhaftere Ergebnis, dass ein Fall nämlich als betrügerisch eingestuft wird, können weiterhin für die beiden Geschlechter abweichen.}

\begin{table}[ht]
  \begin{subfigure}{.48\textwidth}
    \setboolean{TNR}{true} 
    \confusionmatrix{9}{3}{4}{20}{$A$=0}
    \caption{\en{Men}\de{Männer}}
  \end{subfigure}
  \hfill
  \begin{subfigure}{.48\textwidth}
    \setboolean{TNR}{true} 
    \confusionmatrix{6}{3}{3}{15}{$A$=1}
    \caption{\en{Women}\de{Frauen}}
  \end{subfigure}
  \caption{\en{Optimised for Predictive Equality}\de{Optimiert für Predictive Equality}}
  \label{table:confusionmatrices_predicitveequality}
\end{table}

\subsubsection{Balance}\label{ssec:balance}
\en{All previous fairness notions which aim at satisfying separation took binary outputs as a basis. The balance definition uses the predicted probability score instead and compares the average score for both groups per class. This approach seeks to avoid steadily lower outcomes in one group, which might go unnoticed in the binary case, and instead achieve balanced scores for both groups. Depending on the objective it is possible to balance for the positive or negative class~\cite{Kleinberg2016}.}

\de{Alle vorherigen Fairnessdefinitionen, die sich auf das Prinzip der Separierung stützen, haben Prognosen in Form binärer Ausgabewerte verwendet. Bei der Definition Balance werden stattdessen die Wahrscheinlichkeiten zugrunde gelegt, und deren Durchschnittswerte je Klasse für beide Gruppen verglichen. Dieser Ansatz soll verhindern, dass die Prognosen für eine Gruppe systematisch niedriger ausfallen, was im binären Fall eventuell nicht weiter auffallen würde. Stattdessen strebt diese Fairnessdefinition ausgeglichene Ergebnisse auf Basis der gemittelten Wahrscheinlichkeitswerte je Gruppe an. Abhängig vom Anwendungsfall ist es möglich, eine entsprechende Balance für die positive oder für die negative Klasse anzustreben~\cite{Kleinberg2016}.}

\section{\en{The Dilemma}\de{Das Dilemma}}
\en{Presented with all these different fairness definitions it would be convenient to obtain "complete fairness" – one ultimate solution which satisfies all kinds of fairness at the same time. However, there is mathematical tension across the different fairness definitions, and it has been shown that some of them are actually incompatible with each other in realistic scenarios~\cite{Kleinberg2016,Berk2017,Corbett2018}. In this case, optimising for one metric comes with discounts for another. Taking a closer look this seems obvious considering the links between the fairness definitions and the conditional relations within the confusion matrix: The formulas share some of the cell counts, and the cell counts themselves are related to each other (e.g. the sums across the rows, which represent the numbers of true observations, are fixed). 

In public, especially the trade-off between \hyperref[ssec:calibration]{\color{olive}{calibration}} and equal false positive and false negative rates (\hyperref[ssec:equalised_odds]{\color{olive}{equalised odds}}) was much discussed. The debate was initiated by a ML algorithm called the "Correctional Offender Management Profiling for Alternative Sanctions", or COMPAS, which had been developed by the company Northpointe, Inc. The objective of the COMPAS algorithm was to generate an independent, data derived "risk score" for several forms of recidivism. This kind of algorithm is used in the criminal justice sector in the US to support the judge with particular decisions such as granting of bail or parole. The score is of informative character and the final decision is still up to the judge. In May 2016, the investigative journalism website ProPublica focused attention on possible racial biases in the COMPAS algorithm~\cite{Angwin2016}. Its main argument was based on analysis of the data which showed that the results were biased. In particular, the false positive rate for people who were black was significantly higher compared to people who were white. As a result, black people were disproportionately often falsely attributed a high risk of recidivism. Northpointe, on the other hand, responded to the accusations by arguing that the algorithm effectively achieved \hyperref[ssec:predictive_parity]{\color{olive}{predictive parity}} for both groups~\cite{Dieterich2016}. In a nutshell, this ensured that risk scores corresponded to probabilities of reoffending, irrespective of any skin colour. 

From an objective point of view, it can be stated that both parties make valid and reasonable observations of the data. However, the heated public debate revealed that it is unavoidable to precisely define and communicate the desired fairness objective for an application. This decision usually involves arbitration and compromise. For example in the given scenario, calibration and equalised odds could only be mutually satisfied if one of the following conditions was met: Either the base rates of the sensitive subgroups are exactly identical. Or, the outcome classes are perfectly separable which would allow for creating an ideal classifier that achieves perfect accuracy. Unfortunately, both requirements are very unlikely in real world scenarios.}

\de{Angesichts all dieser unterschiedlichen Fairnessdefinitionen ist der Wunsch naheliegend, eine Art von "kompletter Fairness" zu finden – eine ultimative Lösung also, die allen Typen von Fairness gleichermaßen gerecht wird. Manche der Fairnessformeln hängen jedoch über gemeinsame Variablen zusammen, und für einige von ihnen wurde mathematisch nachgewiesen, dass es zumindest unter Praxisbedingungen unmöglich ist, sie parallel zu optimieren~\cite{Kleinberg2016,Berk2017,Corbett2018}. Stattdessen bringen Verbesserungen für die eine Fairnessdefinition immer Verschlechterungen für eine andere mit sich. Wenn man die Zusammenhänge zwischen den Fairnessdefinitionen und die konditionalen Beziehungen innerhalb der Wahrheitsmatrix betrachtet, ist der Grund hierfür leicht nachvollziehbar: Die Formeln verwenden mitunter dieselben Zellwerte, und die Zellen selbst stehen in Beziehung zueinander (z.B. sind die Summen der Zeilen, welche die tatsächlichen Vorkommnisse für die jeweilige Klasse darstellen, fix). 

In der Öffentlichkeit ist vor allem der Kompromiss zwischen kalibrierten Prognosen (\hyperref[ssec:calibration]{\color{olive}{Calibration}}) sowie gleichen Falsch-positiv und Falsch-negativ-Raten (gleichzusetzen mit \hyperref[ssec:equalised_odds]{\color{olive}{Equalised Odds}}) diskutiert worden. Die Debatte wurde durch einen ML-Algorithmus namens "Correctional Offender Management Profiling for Alternative Sanctions", kurz COMPAS, ausgelöst. Dieser war von der Firma Northpointe, Inc. entwickelt worden, und sein Zweck bestand darin, einen unabhängigen, datengestützten "Risikoscore" für verschiedene Formen von Rückfallkriminalität zu ermitteln. Algorithmen dieser Art werden in den USA in der Strafjustiz verwendet, um den Richter oder die Richterin bei bestimmten Entscheidungen wie der Gewährung von Kaution oder Bewährung zu unterstützen. Der Score hat rein informativen Charakter, die finale Entscheidung liegt weiterhin beim Mensch. Im Mai 2016 hat ProPublica, eine Plattform für investigativen Journalismus, mit einem Artikel für Aufsehen gesorgt, der dem COMPAS-Algorithmus rassistische Entscheidungen vorwirft~\cite{Angwin2016}. Das Hauptargument des Berichts stützt sich auf eine Datenanalyse, die Verzerrungen bei den Prognosen feststellt. Insbesondere fiel die Falsch-positiv-Rate für Schwarze deutlich höher aus, als für Weiße. Konkret bedeutet das, dass Schwarzen überproportional häufig Prognosen ausgestellt werden, die zu Unrecht eine erhöhte Rückfälligkeit suggerieren. Northpointe widersprach dem Vorwurf der Diskriminierung mit dem Argument, dass ihr Algorithmus durchaus faire Entscheidungen träfe, indem er \hyperref[ssec:predictive_parity]{\color{olive}{Predictive Parity}} für beide Gruppen erziele: Der Risikoscore gibt also die Wahrscheinlichkeit eines Rückfalls wieder, und das mit gleicher Zuverlässigkeit für beide Gruppen~\cite{Dieterich2016}. 

Objektiv betrachtet kann festgehalten werden, dass die Argumente beider Seiten richtig und berechtigt sind. Die hitzige Debatte allerdings hat deutlich gemacht, dass es für eine KI-Anwendung unumgänglich ist, die angestrebte Art von Fairness vorab festzulegen und zu kommunizieren. Die Wahl bedarf in der Regel Abwägungen und Kompromisse. Im vorliegenden Fall zum Beispiel könnten Calibration und Equalised Odds theoretisch nur dann gemeinsam erreicht werden, wenn eine der beiden folgenden Konditionen eintritt: Entweder, wenn die beiden Basisraten der Untergruppen exakt identisch sind, oder wenn sich die Klassen perfekt separieren lassen. Letzteres würde nämlich das Erstellen eines idealen, fehlerfreien Prognosemodells möglich machen. Leider sind beide Vorausetzungen unter realen Bedingungen sehr unwahrscheinlich.}

\section{\en{\projectname}\de{Der \projectname}}
\en{Based on the limitations explained in the previous section, we conclude that it is crucial to consciously identify the most appropriate fairness definition for every single use case. To support AI stakeholders with this task, we propose the "\projectname", a schema in form of a decision tree which simplifies the selection process. By settling for the desired ethical principles in a formalised way, this schema not only makes identifying the most appropriate fairness definition a straightforward procedure, but it also helps document the underpinning decisions which may serve as deeper explanations to the end user why a specific fairness objective was chosen for the given application.

In this section, we first explain the general intended usage and then deep dive into the key decision points. Finally, we provide some technical specifications and outline how we hope to see this project evolve.}

\de{Aufgrund der im vorherigen Kapitel beschriebenen Einschränkungen sollte beim Einsatz von Künstlicher Intelligenz für jeden Anwendungsfall vorab sorgfältig die passende Definition von Fairness ausgewählt werden. Um KI-Entscheider in dieser Aufgabe zu unterstützen, haben wir den "\projectname" entwickelt: ein Schema in Form eines Entscheidungsbaums, das den Auswahlprozess systematisch vereinfacht. Dabei hilft die Struktur, die ethischen Grundprinzipien für eine Anwendung festzulegen und die jeweiligen Argumente zu dokumentieren. Im Ergebnis formalisiert dieses Tool den Entscheidungsprozess und ermöglicht es so, die Wahl der implementierten Art von Fairness beispielsweise dem Verbraucher detailliert zu begründen. 

In diesem Kapitel beschreiben wir zunächst die allgemeine Anwendung des Tools und erläutern dann die Knotenpunkte des Entscheidungsbaums im Einzelnen. Schließlich erklären wir einige technische Einzelheiten und führen aus, wie sich dieses Projekt weiterentwickeln könnte.}

\subsection{\en{Usage}\de{Anwendung}}
\en{Primarily, the tool consists of the decision tree in \autoref{fig:flow_chart} which formalises the decision process. There are three different types of nodes: The diamonds symbolise decision points, the white boxes stand for actions and the grey boxes with round corners are the fairness definitions. The arrows which connect the nodes represent the possible choices. For increased usability, the schema is also available as interactive 
\href{https://axa-rev-research.github.io/fairness-compass.html}{online tool\footnote{https://axa-rev-research.github.io/fairness-compass.html}}.
In this version, tooltips with extended information, examples and references facilitate navigating the tree. Furthermore, the interactive online tool can be used to document the decision-making process for a specific application. The decision path can be highlighted in the diagram and the reasoning behind each decision can be added in the form of tooltips. In this way, the tool not only serves the AI stakeholders for decision-making but also as means of communication with the users. Due to the general complexity of the topic and the need for context-dependent solutions, we argue that sharing details with the broader audience when specifying fairness for a given use case is the best way forward to maintain confidence in AI systems.}

\de{In erster Linie besteht das Tool aus dem Entscheidungsbaum in \autoref{fig:flow_chart}, der den Auswahlprozess beschreibt. Das Diagramm enthält drei verschiedene Arten von Symbolen: die Rauten stellen die Entscheidungspunkte dar, die weißen Boxen sind Aktionsfelder, und die grauen Boxen mit runden Ecken symbolisieren die jeweiligen Fairnessdefinitionen. Die Pfeile, welche die Symbole verbinden, repräsentieren die Wahlmöglichkeiten. Um die Nutzbarkeit zu vereinfachen ist das Schema auch als interaktives 
\href{https://axa-rev-research.github.io/fairness-compass.html}{Online Tool\footnote{https://axa-rev-research.github.io/fairness-compass.html}} verfügbar.
In dieser Version kann man den Entscheidungsbaum leichter erkunden, da sich Tooltips mit weiterführenden Informationen, Beispielen und Referenzen einblenden lassen. Das interaktive Tool bietet sich außerdem dazu an, den Entscheidungsprozess für einen bestimmten Anwendungsfall hervorzuheben, und die Begründung für jede Entscheidung in Form von Tooltips zu hinterlegen. Auf diese Weise nützt das Tool nicht nur KI-Verantwortlichen bei der Entscheidungsfindung, sondern es eignet sich auch als Mittel, dem Verbraucher diese Entscheidung zu erklären. KI-Systeme fair zu gestalten ist ein vielschichtiges Thema und bedarf kontextbezogener Lösungen. Wir sind daher überzeugt, dass die breitere Öffentlichkeit in die Details einbezogen werden sollte, um das Vertrauen in KI-gestützte Anwendungen langfristig zu stärken.}

\afterpage{%
\begin{figure}[p]
\thispagestyle{empty}
    \vspace*{-3.5cm}
    \makebox[\linewidth]{
        \en{\includegraphics[width=1.50\linewidth]{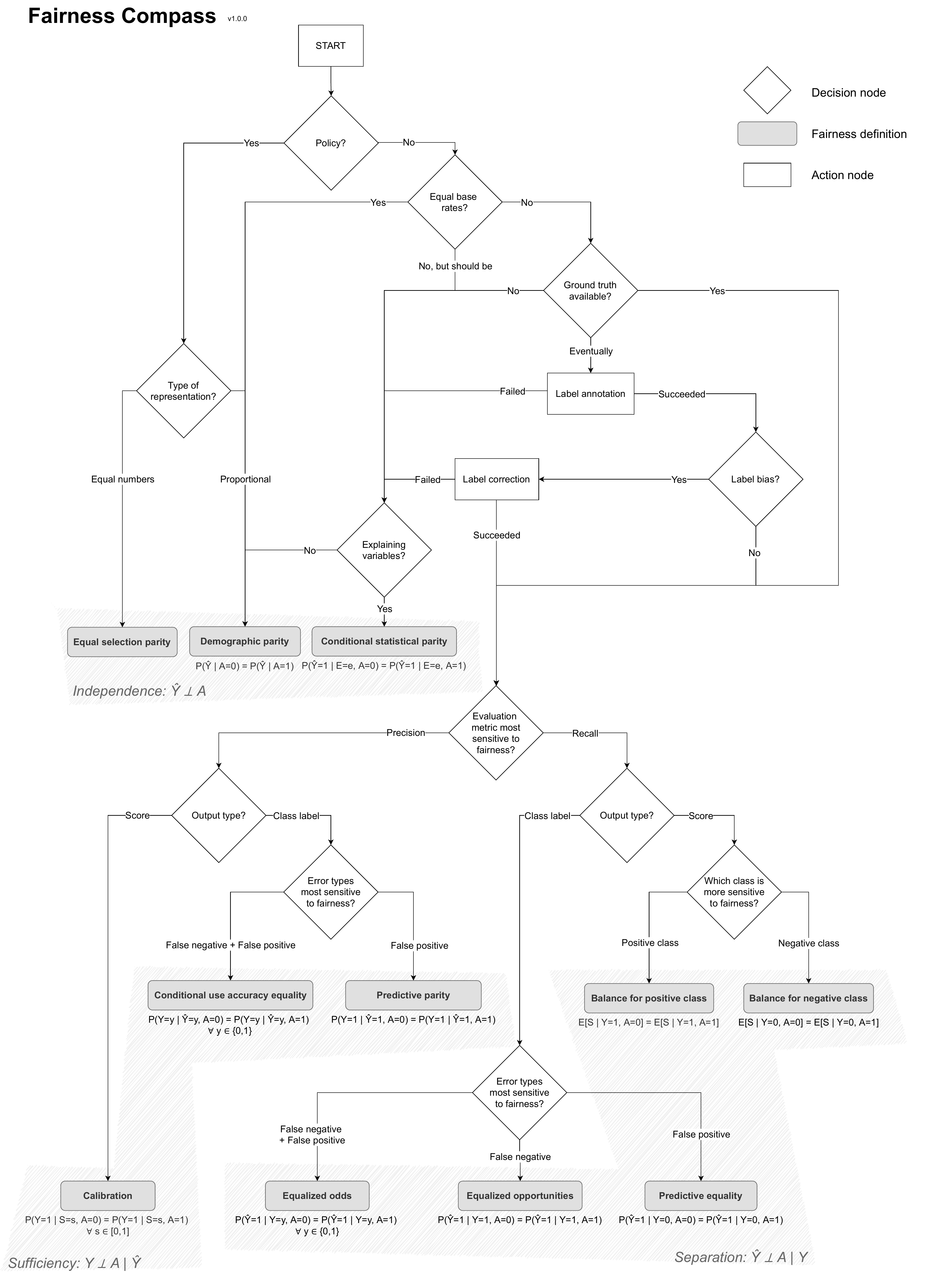}}
        \de{\includegraphics[width=1.50\linewidth]{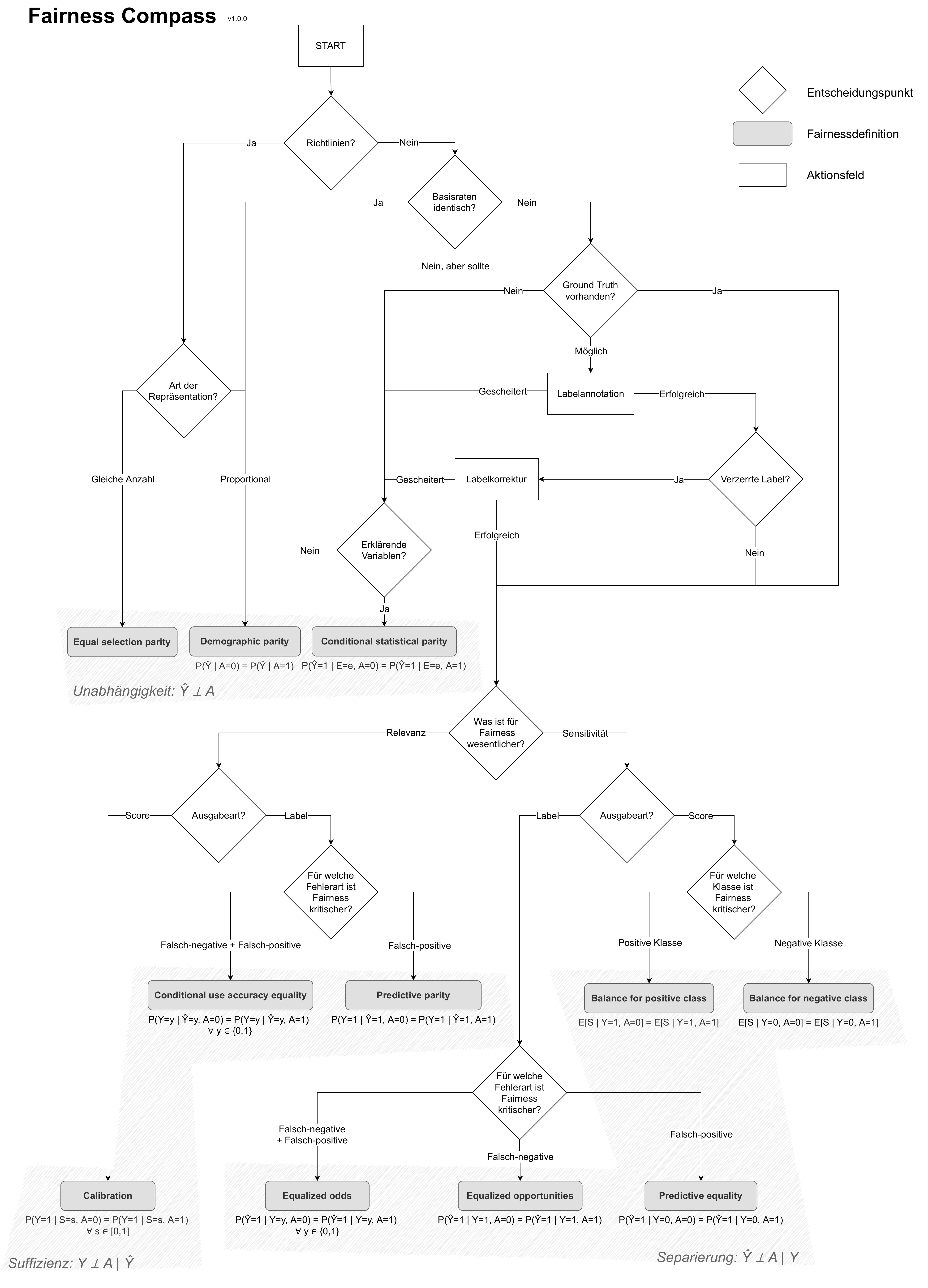}}
    }
    \vspace*{-0.85cm}
    \caption{\en{Proposed schema to structure the complex landscape of fairness\\ definitions for classification. Decision tree also available as 
    \href{https://axa-rev-research.github.io/fairness-compass.html}{online tool}.
    }
    \de{Schema zur Strukturierung des komplexen Angebots von Fairnessdefinitionen für Klassifizierungsprobleme. Auch als 
    \href{https://axa-rev-research.github.io/fairness-compass.html}{Online Tool} abrufbar.
    }}
    \label{fig:flow_chart}
\end{figure}
\clearpage
}

\subsection{\en{Key Decision Nodes}\de{Entscheidungspunkte}}
\en{In the following, we present the major questions we have identified in order to  distinguish between the available fairness definitions. We describe each of them and provide practical examples.}

\de{Im Folgenden präsentieren wir Kernfragen, die wir identifiziert haben, um zwischen den vorhandenen Fairnessdefinitionen zu unterscheiden. Wir beschreiben jeden Punkt im Einzelnen und ergänzen die Ausführungen mit praktischen Beispielen.}

\subsubsection{\en{Policy}\de{Richtlinien}}\label{ssec:policy}
\en{Fairness objectives can go beyond equal treatment of different groups or similar individuals. If the target is to bridge prevailing inequalities by boosting underprivileged groups, affirmative actions or quotas can be valid measures. Such a goal may stem from law, regulation or internal organisational guidelines. This approach rules out any possible causality between the sensitive attribute and the outcome. If the data tells a different story in terms of varying base rates across the subgroups, this is a strong commitment which leads to subordinating the algorithm's accuracy to the policy's overarching goal. In any case, this decision limits the options to fairness definitions which hold the statistical principle of independence (\autoref{ssec:independence}).

For example, many universities aim to improve diversity by accepting more students from disadvantaged backgrounds. Such admission policies acknowledge an equally high academic potential of students from sensitive subgroups and considers their possibly lower level of education rather an injustice in society than a personal shortcoming.}

\de{Zielvorgaben für Fairness können über bloße Gleichbehandlung unterschiedlicher Gruppen oder ähnlicher Individuen hinausgehen. Wenn angestrebt wird, benachteiligte Gruppen direkt zu fördern, um bestehende Ungerechtigkeiten auszugleichen, können Maßnahmen wie "positive Diskriminierung" oder Quoten adäquate Mittel sein. Ein derartiges Ziel kann sich von Gesetzen, Vorgaben von Regulierungsbehörden oder internen Richtlinien einer Organisation ableiten. Dieser Ansatz schließt jeglichen kausalen Zusammenhang zwischen dem sensiblen Attribut und dem Vorhersagewert aus. Falls die vorliegenden Daten in Form von unterschiedlichen Basisraten ein anderes Bild zeichnen, ist das ein starkes Bekenntnis, welches die mathematische Genauigkeit des Algorithmus dieser Überzeugung unterordnet. Wird dieser Entscheidungspunkt bejaht, reduzieren sich die verbleibenden Optionen auf jene Fairnessdefinitionen, die das Prinzip statistischer Unabhängigkeit erfüllen (\autoref{ssec:independence}).

Zum Beispiel haben es sich viele Universitäten zum Ziel gesetzt, ihre Diversität zu erhöhen, indem sie Studenten und Studentinnen aus benachteiligten Verhältnissen bevorzugt Studienplätze anbieten. Eine derartige Zulassungspolitik gesteht diesen Menschen das gleiche akademische Potential wie Studierenden aus privilegierten Kreisen zu, und macht für ihren eventuell niedrigeren Bildungsstand vielmehr gesellschaftliche Missstände als persönliches Versagen verantwortlich.}

\subsubsection{\en{Type of Representation}\de{Art der Repräsentation}}\label{ssec:type_of_representation}
\en{If the decision node about policies from the previous section was answered with yes, particular emphasis is placed on equal representation of the sensitive subgroups. In this case, two different types of representation exist: equal numbers, regardless of the sizes of the subgroups; or proportional equality.

Let's assume a recruitment scenario, for example, where ten women and two men applied for a job. Inviting two women and two men for an interview would satisfy gender fairness based on equal numbers. For proportional equality, it would be necessary to invite five women and one man.}

\de{Wenn die Entscheidung im vorangegangen Punkt zur Beachtung von Richtlinien positiv ausgefallen ist, wird ein besonderer Schwerpunkt auf die ausgewogene Vertretungen der sensiblen Untergruppen gelegt. In diesem Fall gibt es zwei verschiedene Arten von Repräsentation: gleiche Anzahl, unabhängig von den Größen der Untergruppen; oder proportionale Vertretung.

Beispielsweise angenommen, auf eine Stellenausschreibung bewerben sich zehn Frauen und zwei Männer. In Bezug auf das Geschlecht wäre eine Vertretung in gleicher Anzahl gegeben, wenn zwei der Frauen sowie die beiden Männer zum Vorstellungsgespräch eingeladen würden. Um eine proportionale Vertretung zu gewährleisten, müssten fünf Frauen und ein Mann eingeladen werden.}

\subsubsection{\en{Base Rates}\de{Basisraten}}\label{ssec:base_rates}
\begin{figure}[]
    \centering
    \begin{tabular}{ m{6em} m{4em} } 
         $BR=\frac{P}{P+N}$ &
        \includegraphics[height=0.15\textwidth]{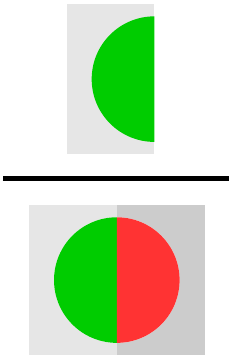}
    \end{tabular}
    \caption{\en{Formula and graphical representation of the (positive) base rate}\de{Formel und grafische Darstellung der (positiven) Basisrate}}
    \label{fig:base_rate}
\end{figure}

\en{Provided there is no policy in place which determines further proceedings, the next crucial question to settle concerns the base rate. This measure was already briefly introduced and it constitutes the proportion of actual positives or actual negatives of the entire data set (recapped in \autoref{fig:base_rate}). Across subgroups, the base rate can be equal, or it can be different. In case of varying rates, it is necessary to decide if the fairness definition should reflect this discrepancy or not. The former is the case when it is legitimate to assume a causal relationship between the group membership and the base rate, and the fairness definition is supposed to take this into account. The latter would be appropriate if there is no rational reason per se to believe that the groups perform differently, and the origin of the different base rates is rather to be found in the data collection process or other data related reasons. Yet another reason to choose equal rates as a basis, even though the data suggest otherwise, is when the discrepancy is considered to originate from historical discrimination. If the fairness objective is meant to make up for such social injustice in the past, assuming equal base rates helps to push the underprivileged group. 

In \cite{Friedler2016}, this question is formalised as two opposing worldviews: The worldview \emph{what you see is what you get (WYSIWYG)} assumes the absence of structural bias in the data. Accordingly, this view supposes that any statistical variation in different groups actually represents deviating base rates which should get explored. On the other hand, the worldview \emph{we're all equal (WAE)} presupposes equal base rates for all groups. Possible deviations are considered as unwanted structural bias that needs to get corrected. 

If base rates are assumed or expected to be equal across subgroups, only fairness definitions which satisfy independence (\autoref{ssec:independence}) remain eligible. Otherwise, if the implemented fairness is to reflect the different base rates, definitions which hold sufficiency or separation (subsections~\ref{ssec:sufficiency} and \ref{ssec:separation}) are to be considered.

In a medical scenario, the base rates for women and men to suffer from diabetes are equal, while 99\% of breast cancer occurs in women. A fair diagnostic application should acknowledge this discrepancy. In a college admission scenario, however, different base rates in admission exams across different ethnic groups could be attributed to unequal opportunities. If the declared objective of the fairness definition is to correct such social injustice, choosing equal base rates can be a suitable approach.}

\de{Vorausgesetzt es sind keine Richtlinien zu befolgen, die eine repräsentative Verteilung vorschreiben, dann betrifft die nächste Frage die Basisrate. Diese statistische Kennzahl wurde bereits eingeführt und beschreibt den Anteil der tatsächlich positiven oder negativen Fälle vom gesamten Datensatz (wiederholt in \autoref{fig:base_rate}). Zwischen den Untergruppen kann die Basisrate gleich ausfallen, oder sie kann abweichen. Wenn die Raten abweichen muss entschieden werden, ob die Fairnessdefinition diesen Unterschied reflektieren soll, oder nicht. Der erste Fall ist gegeben, wenn es einen berechtigten Grund zur Annahme gibt, dass ein Kausalzusammenhang zwischen der Gruppenzugehörigkeit und der Basisrate besteht, und die Fairnessdefinition diesen Effekt berücksichtigen soll. Der zweite Fall gilt, wenn es keine rationale Erklärung gibt, warum die beiden Gruppen grundsätzlich unterschiedliche Ergebnisse liefern sollten. Dann wird die Ursache vielmehr beim Datenerhebungsverfahren vermutet, oder auf andere datenbezogene Aspekte zurückgeführt. Ein weiterer Anlass, von gleichen Basisraten auszugehen, obwohl die Daten andere Rückschlüsse zulassen, liegt vor, wenn für die Abweichung soziale Diskriminierung in der Vergangenheit verantwortlich gemacht wird. Soll die gewünschte Fairnessdefinition eine derartige historische Ungerechtigkeit ausgleichen helfen, stärkt es die Position der unterprivilegierten Gruppe, wenn trotzallem von gleichen Basisraten ausgegangen wird. 

In \cite{Friedler2016} wird diese Frage als zwei gegensätzliche Weltanschauungen definiert: Die Weltanschauung \emph{What you see is what you get (WYSIWYG)} nimmt an, dass keine strukturellen Verzerrungen in den Daten existieren. Entsprechend geht diese Theorie davon aus, dass statistische Abweichungen in den Basisraten für das Prognosemodell von Relevanz sind und berücksichtigt werden müssen. Dem gegenüber steht die Weltanschauung \emph{We're all equal (WAE)}, welche grundsätzlich gleiche Basisraten für alle Gruppen vermutet. Etwaige Abweichungen werden als unerwünschte, strukturelle Verzerrungen interpretiert, die es zu korrigieren gilt. 

Wenn davon ausgegangen wird, dass die Basisraten aller Untergruppen identisch sind, kommen als Fairnessdefinitionen nur jene in Frage, die das Prinzip der unbedingten statistischen Unabhängigkeit erfüllen (\autoref{ssec:independence}). Anderenfalls, wenn die zu implementierende Fairness die unterschiedlichen Basisraten widerspiegeln soll, kommen Fairnessdefinitionen, die Suffizienz oder Separierung erfüllen (Unterabschnitte~\ref{ssec:sufficiency} und \ref{ssec:separation}) in Betracht.

In einem medizinischen Kontext zum Beispiel kann festgestellt werden, dass die Basisraten für Männer und Frauen, die an Diabetes leiden, etwa gleich sind, wohingegen Brustkrebs in 99\% der Fällen bei Frauen auftritt. Eine faire Diagnoseanwendung sollte diesen Unterschied berücksichtigen. Bei Zulassungstests im Rahmen einer Studienplatzvergabe hingegen könnten unterschiedliche Basisraten bei verschiedenen ethnischen Gruppen auf Chancenungleichheiten zurückzuführen sein. Besteht der Anspruch, soziale Ungerechtigkeit auszugleichen, kann hier der passende Ansatz sein, trotzdem gleiche Basisraten anzusetzen.}

\subsubsection{Ground Truth}\label{ssec:ground_truth}
\en{Machine learning algorithms are trained by example. The assumption is that the labels of the training data represent the true output, they constitute the supposed ground truth. As the labels serve as reference to estimate the model's accuracy, but also to satisfy a fairness metric when this one is conditioning on the label, the availability of a reliable ground truth makes a significant difference.

Depending on the scenario, the ground truth may not exist or it may exist but not be available. When the correct outcome can be observed, the ground truth exists, and when the labels result from objective measurements or describe indisputable facts, the truth is easily accessible. In other cases, the correct outcome may not be directly measurable, but it is still unambiguously observable by humans who can perform annotation tasks with sufficient diligence to produce reliable labels. Sometimes, the ground truth does not exist. In such a scenario, labels are only inferred and represent subjective human decisions based on experience, and they may contain human bias.

When the ground truth is not available, and it cannot be produced in a trustworthy way neither, it is not recommended to select fairness definitions which rely on the true output value as is the case for the separation principle (\autoref{ssec:separation}). Under these conditions, it is rather advisable to choose from fairness definitions which satisfy independence (\autoref{ssec:independence}) and do not condition on the training label. 

In a medical scenario, it is possible to conclusively clarify if a tumour is benign or malignant by taking a biopsy and performing a laboratory examination. Hence, the ground truth is available. In an image recognition application which helps classify animals, humans can make training data of good quality available by manually labelling the different species. When predicting recidivism, the ground truth is not immediately available since a possible new criminal offence would take place in the future and may not even be caught.}

\de{ML-Algorithmen lernen von Beispielen. Dabei wird vorausgesetzt, dass die Label der Trainingsdaten den wahren Ausgabewert repräsentieren – sie stellen die sogenannte \textit{Ground Truth} dar. Da diese Label auch zur Bewertung der Genauigkeit des Prognosemodells herangezogen werden, und sogar eine Referenz für das Fairnessmaß abgeben, falls die gewählte Fairnessdefinition das Label bedingt, spielt die Präsenz einer verlässlichen Ground Truth eine erhebliche Rolle.

Abhängig vom Anwendungsfall kann es allerdings vorkommen, dass keine Ground Truth vorhanden ist, oder dass diese zwar existiert, aber nicht direkt zugänglich ist. Wenn sich das wahre Ergebnis beobachten lässt, besteht eine Ground Truth, und wenn die Label zum Beispiel objektive Messungen oder eindeutige Fakten darstellen, ist der Zugriff meistens nicht weiter schwierig. In anderen Fällen kann das korrekte Ergebnis zwar nicht maschinell erfasst werden, aber es kann von Menschen dennoch eindeutig beobachtet werden, die per sorgfältiger, manueller Zuordnung ausreichend verlässliche Label erzeugen können. Manchmal jedoch ist auch keine Ground Truth vorhanden. Unter diesen Umständen werden die Label abgeleitet und basieren auf menschlichen Entscheidungen, die auf subjektiver Erfahrung beruhen. Solche Label sind möglicherweise vorurteilbehaftet. 

Ist keine Ground Truth vorhanden, und lässt sich diese auch nicht auf verläss\-liche Weise erschließen, dann reduzieren sich die Auswahlmöglichkeiten. Fairnessdefinitionen, die das Prinzip der Separierung (\autoref{ssec:separation}) erfüllen und die wahre Klasse bedingen, sind in diesem Fall nicht zu empfehlen. Stattdessen kommt eher eine Fairnessdefinition in Frage, die das Prinzip der Unabhängigkeit erfüllt (\autoref{ssec:independence}) und keine Trainingslabel voraussetzt. 

In einem medizinischen Szenario kann für einen Tumor mithilfe einer Biopsie und anschließender Laboruntersuchung abschließend geklärt werden, ob er gut- oder bösartig ist. Die Ground Truth ist also verfügbar. In einer Bilderkennungssoftware wiederum, die Tierarten spezifizieren soll, können menschliche Experten Fotos manuell annotieren und so Trainingsdaten von guter Qualität erzeugen. Soll allerdings die Rückfallkriminalität vorhergesagt werden, ist keine Ground Truth unmittelbar verfügbar, da mögliche neue Straftaten erst in der Zukunft stattfinden, und außerdem gar nicht jedes Verbrechen aktenkundig wird.}

\subsubsection{\en{Explaining Variables}\de{Erklärende Variablen}}\label{ssec:explaining_variables}
\en{The data may contain variables which are considered legitimate sources of discrepancy. If some kind of inequality between the groups can be shown to stem from those variables, this sort of discrimination can be considered explainable and accepted~\cite{Kamiran2013}.

Let's suppose salary ranges are to be estimated for job applicants. However, the training data show that one group works fewer hours on average than the other. In this case, a variable \emph{working\_hours} could serve as an explaining variable.}

\de{Unter Umständen gibt es Attribute, die als legitime Quelle für Unterschiede in den Daten betrachtet werden können. Wenn aufgezeigt werden kann, dass eine Verzerrung in den Daten zwischen den Untergruppen auf diese Variablen zurückzuführen ist, kann dieser Unterscheid begründet und die Abweichung akzeptiert werden~\cite{Kamiran2013}.

Angenommen es sollen Gehaltsklassen für Jobbewerber geschätzt werden. Im vorliegenden Datensatz arbeitet eine Gruppe aber im Durchschnitt weniger Arbeitsstunden als die andere. Dann könnte ein Attribut \emph{working\_hours} als erklärende Variable dienen.}

\subsubsection{\en{Label Bias}\de{Verzerrte Label}}\label{ssec:label_bias}
\en{When no ground truth exists and the available labels are based on decisions which were inferred by humans, they may contain human bias. As the labels serve as reference to estimate the model's accuracy but also to satisfy a fairness metric when this one is based on classification rates, it is crucial to mitigate this potential source of bias, possibly using a label correction framework~\cite{Wick2019, Jiang2019}. If such action does not yield satisfying results, the ground truth is missing and therefore the same reasoning as before applies: It is not recommendable to use fairness definitions which condition on the true output value but rather choose from the ones which hold independence (\autoref{ssec:independence}).

For example, if software is to learn to describe photos with words, then humans generate the training data by tagging sample images. This task allows for a certain degree of creative freedom, for example in the selection of objects or their description. Especially if this activity is performed by only a small group of people, the training data may include human bias.}

\de{Wenn die Ground Truth nicht direkt zugänglich ist, und die vorhandenen Label von Menschen manuell zugeordnet wurden, besteht das Risiko, dass die so erhobenen Daten gewisse Verzerrungen enthalten. Die Label werden zur Bestimmung der Genauigkeit des Prognosemodells herangezogen, und spielen auch für die Bewertung der Fairness eine Rolle, falls deren Definition die Label berücksichtigt. Daher ist es entscheidend, möglichen Quellen von Verzerrung entgegenzuwirken, zum Beispiel mit einem Labelkorrektur-Framework~\cite{Wick2019, Jiang2019}. Sollte dieser Vorgang keine zufriedenstellenden Ergebnisse liefern, ist keine belastbare Ground Truth vorhanden, und es gilt die gleiche Argumentation wie zuvor: Es ist nicht ratsam, eine Fairnessdefinition zu verwenden, welche sich auf den wahren Ausgabewerte stützt, sondern besser eine Definition zu wählen, die das Prinzip der Unabhängigkeit erfüllt (\autoref{ssec:independence}).

Wenn zum Beispiel eine Software lernen soll, Fotos mit Worten zu beschreiben, dann erzeugen Menschen die zugehörigen Trainingsdaten, indem sie Beispielbilder verschlagworten. Diese Aufgabe lässt einen gewissen kreativen Freiraum zu, etwa durch die Auswahl der Objekte, oder deren Bezeichnung. Insbesondere wenn diese Tätigkeit nur von einer kleinen Gruppe Menschen ausgeübt wird, können sich in den Trainingsdaten Verzerrungen niederschlagen.}

\subsubsection{\en{Precision and Recall}\de{Relevanz und Sensitivität}}\label{ssec:evaluation_metric}
\begin{figure}[h]
\centering
  \begin{subfigure}{.49\textwidth}
  \centering
      \begin{tabular}{ m{7em} m{3em} } 
         $PPV=\frac{TP}{TP+FP}$ &
        \includegraphics[height=0.3\textwidth]{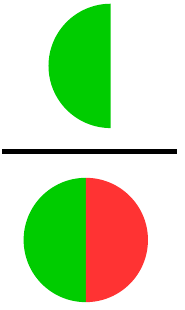}
    \end{tabular}
    \caption{\en{Precision}\de{Relevanz}}
  \end{subfigure}
  \begin{subfigure}{.49\textwidth}
      \centering
      \begin{tabular}{ m{6em} m{3em} } 
         $TPR=\frac{TP}{P}$ &
        \includegraphics[height=0.3\textwidth]{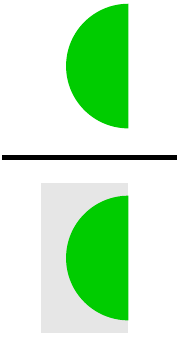}
    \end{tabular}
    \caption{\en{Recall}\de{Sensitivität}}
  \end{subfigure}
  
  \caption{\en{Formulas and graphical representations of precision and recall}\de{Formeln und symbolische Darstellung beider Metriken}}
  \label{fig:precision_recall}
\end{figure}

\en{After concluding that some sort of reliable ground truth is available, a well-known problem in machine learning needs to be tackled: the trade-off between precision and recall. Precision describes the fraction of positively predicted instances which were actually positive, previously introduced as positive predictive value (PPV). Recall is the fraction of actual positive instances which were correctly identified as such, also defined as true positive rate (TPR) earlier in this document. \autoref{fig:precision_recall} brings back the two formulas in a graphical representation. The question to be addressed here is which of the two metrics is more sensitive to fairness in the given use case. A general rule is that when the consequences of a positive prediction have a negative, punitive impact on the individual, the emphasis with respect to fairness often is on precision. When the result is rather beneficial in the sense that the individuals are provided help they would otherwise forgo, fairness is often more sensitive to recall. The answer to this decision node also determines to which of the two remaining categories the ultimate fairness definition will belong to: If the focus is on equal precision rates for the sensitive subgroups, the final definition will condition on the predicted output and therefore hold sufficiency (\autoref{ssec:sufficiency}). Otherwise, if the focus is on equal recall rates, the resulting fairness definition will condition on the true output and satisfy separation (\autoref{ssec:separation}).

In a fraud detection scenario where insurance claims are to be investigated it could be considered fair to limit the number of wrongly suspected cases and therefore maximise precision at equally high level for all subgroups. In a loan approval scenario, the focus regarding fairness could be on recall, that is, approving an equally high level of loans to creditworthy applicants across all groups.}

\de{Wenn feststeht, dass eine ausreichend verlässliche Ground Truth vorhanden ist, muss als nächstes über ein bekanntes Problem beim Maschinellen Lernen entschieden werden: der Kompromiss zwischen Relevanz und Sensitivität. Relevanz (\textit{precision}) beschreibt den Anteil der positiven Vorhersagen, der korrekt ist – zuvor eingeführt als positiver Vorhersagewert (PPV). Sensitivität (\textit{recall}) beschreibt den Anteil der tatsächlich positiven Fälle, der korrekt vorhergesagt wurde – in diesem Bericht auch als Richtig-Positiv-Rate (TPR) bezeichnet. In \autoref{fig:precision_recall} stellen wir beide Kennzahlen erneut mathematisch und symbolisch dar. Zu klären ist nun, welches der beiden Maße für den gegebenen Anwendungsfall in Bezug auf die Fairness eine kritischere Rolle spielt. Als Faustregel gilt, dass wenn die Konsequenzen für den oder die Betroffene im Zweifelsfall eine negative, strafende Auswirkung haben, dann der Schwerpunkt bezüglich Fairness auf der Relevanz liegen sollte. Ist das Ergebnis im besten Fall vielmehr vorteilhaft, im Sinne von Unterstützung, auf die die Person sonst verzichten müsste, dann ist oft der Aspekt der Sensitivität wesentlicher in puncto Fairness. Die Antwort auf diese Frage legt fest, aus welcher der verbleibenden beiden Kategorien die endgültige Fairnessdefinition stammen wird: Wenn der Schwerpunkt auf gleichen Relevanzraten für beide Untergruppen liegt, dann wird die Fairnessdefinition auf der vorhergesagten Klasse beruhen, und folglich dem Prinzip der Suffizienz genügen (\autoref{ssec:sufficiency}). Anderenfalls, wenn der Fokus auf der Sensitivität liegt, wird die resultierende Fairnessdefinition auf der wahren Klasse basieren und das Separierungs-Prinzip erfüllen (\autoref{ssec:separation}).

In einem Anwendungsfall, bei dem es um Betrugserkennung bei Versicherungsfällen geht, könnte man es als oberstes Fairnessziel betrachten, die Zahl der zu Unrecht als betrügerisch eingeordneten Fälle zu minimieren und für beide Untergruppen die Relevanzraten gleich niedrig zu halten. In einem Kreditvergabeszenario hingegen könnte der Fokus in Bezug auf Fairness bei der Sensitivität liegen. Das hieße, kreditwürdigen Bewerbern und Bewerberinnen aus beiden Untergruppen sollten ihre Anträge zu gleich hohen Raten bewilligt werden.}

\subsubsection{\en{Output Type}\de{Ausgabearten}}\label{ssec:output_type}
\en{A more practical than ethical question, but nonetheless relevant to determining the ultimate fairness definition, is that of the desired output type. A score is a continuous value, often between 0 and 1, which can represent the probability for the positive class to be true. When the output is a label instead, the result is an unambiguous decision for one of the classes.

For example in a loan approval scenario, a score is often preferred because the value of the score leaves the "human in the loop" some room for interpretation. However, when the result is automatically processed, for example in an online marketing scenario, the class label may be the preferred output type.}

\de{Eine mehr praktische als ethische Frage, aber dennoch relevant um die finale Fairnessdefinition zu wählen, ist jene nach der gewünschten Ausgabeart. Ein Score ist ein kontinuierlicher Wert. Oftmals liegt er zwischen 0 und 1, und beschreibt dann die Wahrscheinlichkeit, dass der gegebene Fall der positiven Klasse angehört. Ist die Ausgabeart stattdessen ein Label, entspricht das Ergebnis eindeutig einer der beiden Klassen.

In einem Kreditvergabeszenario wird oft ein Score bevorzugt, weil sein Wert mehr Interpretationsspielraum für den Mensch lässt, der die finale Entscheidung trifft. Wird das Ergebnis allerdings automatisch weiterverarbeitet, zum Beispiel in einem Online-Marketing Szenario, könnten Klassenlabel als Ausgabeart sinnvoller sein.}

\subsubsection{\en{Error Types}\de{Fehlerarten}}\label{ssec:error_types}
\begin{figure}[h]
\centering
  \begin{subfigure}{.49\textwidth}
  \centering
      \begin{tabular}{ m{6em} m{3em} } 
         $FPR=\frac{FP}{N}$ &
        \includegraphics[height=0.3\textwidth]{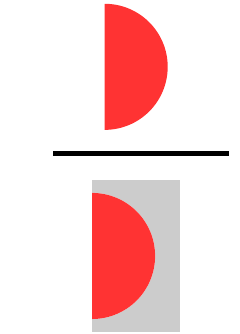}
    \end{tabular}
    \caption{\en{False positive rate}\de{Falsch-positiv-Rate}}
  \end{subfigure}
  \begin{subfigure}{.49\textwidth}
      \centering
      \begin{tabular}{ m{6em} m{3em} } 
         $FNR=\frac{FN}{P}$ &
        \includegraphics[height=0.3\textwidth]{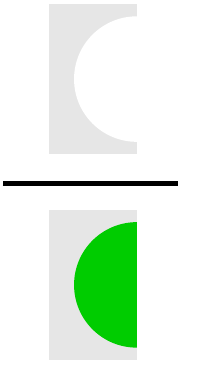}
    \end{tabular}
    \caption{\en{False negative rate}\de{Falsch-negativ-Rate}}
  \end{subfigure}
  
  \caption{\en{Formulas and graphical representations of error rates}\de{Formeln und symbolische Darstellungen der Fehlerarten}}
  \label{fig:false_positive_and_false_negative_rates}
\end{figure}

\en{Eventually, the final decision depends on which error types are considered most sensitive to fairness for the use case. The different error types to take into account are the false positive and the false negative rate (as introduced earlier and recapped in \autoref{fig:false_positive_and_false_negative_rates}). Both represent measures of misclassification, but based on the use case, one error type may be more sensitive to fairness than the other. Generally, the goal for high-risk applications is to keep positive and negative classification rates equal for all groups. For low-risk applications the fairness objective could be weakened by accepting a manageable degree of extra risk in order to increase utility of the metric~\cite{Hardt2016}. For better clarity, it may be helpful to enhance the confusion matrix (see \autoref{ssec:statistical_measures}) by the expected benefits and harms in order to visualise the consequences of correct or false classification scenarios and weight the error types accordingly.

In an online marketing scenario where a job offer is supposed to be shown to men and women of relevant profiles, differences in false positive rates (showing the ad to people who are not eligible) across the groups may be acceptable as long as the fractions of people with relevant profiles are equal. On the other hand, in a face recognition application both error types should be equally low for all sorts of skin types.}

\de{Die letzte Entscheidung betrifft schließlich die Frage, welche Fehlerart im gegebenen Anwendungsfall eine höhere Priorität bezüglich Fairness hat. Die unterschiedlichen zu berücksichtigenden Fehlerarten sind die Falsch-positiv und die Falsch-negativ-Rate (wie bereits etwas früher eingeführt und in \autoref{fig:false_positive_and_false_negative_rates} rekapituliert). Beide Kennzahlen bemessen Fehlklassifikation, abhängig vom Einsatzgebiet kann eine Fehlerart allerdings eine bedeutsamere Rolle beim Erreichen von Fairness haben, als die andere. Generell gilt, dass für Hochrisiko-Anwendungen sowohl die Falsch-positiv als auch die Falsch-negativ-Rate für alle Gruppen auf gleichem Niveau gehalten werden sollte. Für weniger sicherheitskritische Anwendungen könnte das Fairnessziel zugunsten erhöhter technischer Flexibilität etwas abgeschwächt werden, indem ein überschaubares zusätzliches Risiko in Kauf genommen wird~\cite{Hardt2016}. Um in dieser Angelegenheit einen besseren Überblick zu haben, kann es nützlich sein, die Wahrheitsmatrix (siehe \autoref{ssec:statistical_measures}) um eine Beschreibung der Ereignisse im Fehlerfall zu ergänzen. So lassen sich die Konsequenzen einer korrekten oder inkorrekten Einordnung vor Augen führen und entsprechend gewichten.

In einem Online-Marketing Szenario, wo eine Stellenanzeige Männer und Frauen mit relevanten Profilen eingeblendet werden soll, mögen Unterschiede in der Falsch-positiv-Rate (die Anzeige also Menschen zu zeigen, die eigentlich nicht für die Stelle in Frage kommen) zwischen den Untergruppen verkraftbar sein, solange die Anteile der Leute mit relevanten Profilen gleichermaßen hoch sind. Bei einer Gesichtserkennungssoftware andererseits sollten beide Fehlerarten für sämtliche Hauttypen gleich niedrig sein.}

\subsection{\en{Sample Application}\de{Beispielanwendung}}
\en{We apply the "\projectname" on our running sample scenario about fraud detection in insurance claims. Thanks to our online tool, it becomes a straightforward and transparent task to provide a possible chain of arguments (see 
\href{https://axa-rev-research.github.io/fairness-fraud-study.html}{online resource\footnote{https://axa-rev-research.github.io/fairness-fraud-study.html}}). Please note that this example remains a thought experiment. For the same scenario, other considerations with different outcomes are possible. The purpose of the "\projectname" is not to impose one solution but to assist with the decision making and to underpin the result.}

\de{Wir testen den "\projectname" für unser wiederkehrendes Beispiel zur Betrugserkennung bei Versicherungsfällen. Mit unserem interaktiven Tool lässt sich die Entscheidungsfindung für die Wahl einer Fairnessdefinition einfach und transparent darstellen (siehe 
\href{https://axa-rev-research.github.io/fairness-fraud-study.html}{Online-Tool\footnote{https://axa-rev-research.github.io/fairness-fraud-study.html}}). Wohlgemerkt handelt es sich bei diesem Beispiel um ein Gedankenexperiment. Für dasselbe Szenario sind verschiedene Argumente denkbar, die zu einem anderen Ergebnis führen. Der Zweck des "\projectname" ist nicht, eine Lösung vorzuschreiben, sondern vielmehr die Entscheidungsfindung zu strukturieren und das Ergebnis argumentativ zu rechtfertigen.}

\subsection{\en{Future Development}\de{Weitere Entwicklung}}
\en{Research in fair machine learning is constantly advancing, new types of fairness definitions may evolve and the general debate on fairness will move on. To anticipate those future developments, our technical architecture puts easy expandability and adaptability at the centre. The online tool was realised using the free online diagram software \href{https://www.diagrams.net/}{diagrams.net}. We used it to design the tree and to publish it online. The schema is encoded in XML which allows the use of version control to track modifications and enhancements. We further developed a plugin for diagrams.net to extend its scope of functions for the interactive features we described above. We also made the \href{https://axa-rev-research.github.io/fairness-compass/src/main/webapp/plugins/props.js}{source code\footnote{https://axa-rev-research.github.io/fairness-compass/src/main/webapp/plugins/props.js}} of this plugin publicly available.}

\de{Die Forschung im Bereich KI und Fairness schreitet stetig voran. Wahrscheinlich werden neue Fairnessdefinitionen entwickelt werden. Die allgemeine Fairness-Debatte wird sicherlich fortgesetzt werden, und die Gesellschaft wird konkretere Erwartungen formulieren, wie faire Entscheidungen in bestimmten Anwendungsbereiche auszusehen haben. Um diese zukünftigen Entwicklungen zu berücksichtigen, haben wir bei unserer technischen Architektur den Schwerpunkt auf Erweiterbarkeit und Anpassbarkeit gelegt. Das Online-Tool wurde mit der kostenlosen Online-Diagrammsoftware \href{https://www.diagrams.net/}{diagrams.net} realisiert. Wir haben damit den Entscheidungsbaum entworfen und online veröffentlicht. Das Schema ist im XML-Format gespeichert, was eine Versionierung und Nachverfolgung von Änderungen und Erweiterungen möglich macht. Wir haben außerdem ein Plug-In für diagrams.net implementiert, das dessen Funktionsumfang um die oben beschriebenen interaktiven Features erweitert. Den \href{https://axa-rev-research.github.io/fairness-compass/src/main/webapp/plugins/props.js}{Quellcode\footnote{https://axa-rev-research.github.io/fairness-compass/src/main/webapp/plugins/props.js}} für dieses Plugin haben wir ebenfalls online zur Verfügung gestellt.}

\section{\en{Conclusion}\de{Schlussfolgerungen}}
\en{This document seeks to explain in a comprehensible way the problem of bias in AI, and why there is no silver bullet to overcome it. We provide background information on a various list of fairness definitions for classification problems in machine learning and illustrate their different properties by example. As a practical approach for better orientation in the complex landscape of fairness definitions, we further propose the "\projectname", a decision tree which outputs the best suited option for a given use case after settling a few crucial questions on the desired type of fairness. This tool also helps document the reasoning behind the selection process which contributes to more transparency and potentially provides better understanding and increased trust among the affected users.

We would like to point out that the presented diagram is certainly not the last word on the subject. Research in fair machine learning is constantly advancing, new types of fairness definitions may evolve and the general debate on fairness will move on. Therefore, we consider this work as first step towards structuring the complex landscape of fairness definitions. We would be happy to see this project help illustrate the decision making in particular application scenarios but also serve as a basis for fundamental discussions and further refinements in the course of implementing fair machine learning. Eventually, we hope that our proposal makes a useful contribution to a smooth implementation of fair machine learning in real world applications.}

\de{Dieser Bericht behandelt das Problem von Verzerrungen in Daten und KI-Anwendungen. Wir erklären seine Ursachen und Konsequenzen, und führen zudem aus, warum es für die Lösung keinen Königsweg gibt. Zur allgemeinen Übersicht stellen wir die verfügbaren Fairnessdefinitionen für Klassifizierungsprobleme vor, und erläutern deren unterschiedlichen Eigenschaften anhand von Beispielen. Als Orientierungshilfe und praktischen Lösungsansatz präsentieren wir den "\projectname" – ein Tool in Form eines Entscheidungsbaums, das einige wesentliche Fragen zur gewünschten Art von Fairness abfragt, und dann auf Grundlage der Antworten  für einen konkreten Anwendungsfall die am besten passende Option liefert. Dieses Tool ist auch nützlich, um die Argumente bei der Entscheidungsfindung zu dokumentieren. Menschen, die von der Entscheidung eines KI-Systems betroffen sind, können so einfacher nachvollziehen, warum die vorliegende Art von Fairness implementiert wurde. Erhöhte Transparenz in dieser Form kann dazu beitragen, das Vertrauen in KI-Anwendungen zu stärken.

Wir möchten betonen, dass das hier vorgestellte Schema wohl nicht das letzte Wort zu diesem Thema sein wird. Die Forschung wird neue Erkenntnisse liefern, und die gesellschaftlichen Erwartungen werden sich konkretisieren. Wir verstehen unsere Arbeit daher als ersten Schritt, das komplexe aktuelle Angebot von Fairnessdefinitionen zu strukturieren. Wir würden uns freuen, wenn dieses Projekt zu fundierten Entscheidungen in konkreten Anwendungsszenarien beiträgt, und hoffen weiterhin, dass es als Basis für grundsätzliche Diskussionen dient, und so einen nützlichen Beitrag für die Implementierung von mehr Fairness in realen KI-Systemen leistet.}

\section*{\en{Acknowledgements}\de{Danksagungen}}
\en{We thank Jonathan Aigrain for helpful discussion and valuable feedback on this document. Many thanks also to George Woodman for the proofreading.}
\de{Wir bedanken uns bei Jonathan Aigrain für die anregenden Diskussionen und das  konstruktive Feedback zu diesem Dokument.}

\bibliographystyle{unsrt}
\bibliography{bibliography}

\end{document}

%% file: macros/confusion_matrices.tex
\newcommand{\highlight}[2]{
  \ifthenelse{\boolean{#2}}{\textbf{#1}}{#1}
}

\newcommand{\baserate}[4]{
  \fpeval{round((#1+#2)/(#1+#2+#3+#4),2)}
}

\newcommand{\negativerate}[4]{
  \fpeval{round((#2+#4)/(#1+#2+#3+#4),2)}
}

\newcommand{\falsedetectionrate}[4]{
  \fpeval{round((#3)/(#1+#3),2)}
}

\newcommand{\falseomissionrate}[4]{
  \fpeval{round((#2)/(#2+#4),2)}
}

\newcommand{\truepositiverate}[4]{
  \fpeval{round((#1)/(#1+#2),2)}
}

\newcommand{\truenegativerate}[4]{
  \fpeval{round((#4)/(#3+#4),2)}
}

\newcommand{\confusionmatrix}[5]{%

\centering
\arrayrulecolor{black}
\renewcommand\arraystretch{2}
\resizebox{\textwidth}{!}{
\begin{tabular}{|c|c|c|c|c|}
\hhline{~~--~}
\multicolumn{2}{c|}{#5}                                                                                                                                                                  & \multicolumn{2}{c|}{{\cellcolor[rgb]{0.937,0.937,0.937}}\en{Predicted}\de{Prognose}}                                          & \multicolumn{1}{l}{}       \\ 
\hhline{~~---|}
\multicolumn{1}{c}{}                                                                                                                       &                                            & {\cellcolor[rgb]{0.937,0.937,0.937}}$\widehat{Y}$=1  & {\cellcolor[rgb]{0.937,0.937,0.937}}$\widehat{Y}$=0  & \highlight{BR=\baserate{#1}{#2}{#3}{#4}}{BR}               \\ 
\hline
{\cellcolor[rgb]{0.937,0.937,0.937}}                                                                                                       & {\cellcolor[rgb]{0.937,0.937,0.937}}$Y$=1  & \multicolumn{1}{c|}{#1}                             & \multicolumn{1}{c|}{#2}                             & \multicolumn{1}{l|}{\highlight{TPR=\truepositiverate{#1}{#2}{#3}{#4}}{TPR}}  \\ 
\hhline{|>{\arrayrulecolor[rgb]{0.937,0.937,0.937}}->{\arrayrulecolor{black}}----|}
\multirow{-2}{*}{\cellcolor[rgb]{0.937,0.937,0.937}\rotatebox[origin=c]{90}{\en{True}\de{Wahr}} } & {\cellcolor[rgb]{0.937,0.937,0.937}}$Y$=0  & #3                                                  & #4                                                  & \highlight{TNR=\truenegativerate{#1}{#2}{#3}{#4}}{TNR}                     \\ 
\hline
\multicolumn{1}{l}{}                                                                                                                       & \multicolumn{1}{l|}{}                      & \highlight{FDR=\falsedetectionrate{#1}{#2}{#3}{#4}}{FDR}                                                 & \highlight{FOR=\falseomissionrate{#1}{#2}{#3}{#4}}{FOR} & \multicolumn{1}{l}{}       \\
\cline{3-4}
\multicolumn{1}{l}{}                                                                                                                       & \multicolumn{1}{l|}{}                      & PR=\fpeval{round((#1+#3)/(#1+#2+#3+#4),2)}                                                 & \highlight{NR=\negativerate{#1}{#2}{#3}{#4}}{NR} & \multicolumn{1}{l}{}       \\
\cline{3-4}
\end{tabular}}
}

\newcommand{\confusionmatrixterms}{
\centering
\renewcommand\arraystretch{2}
\renewcommand\tabcolsep{10pt}
\arrayrulecolor{black}
\begin{tabular}{|c|c|c|c|} 
\hhline{~~--|}
\multicolumn{1}{c}{}                                                             &                                                  & \multicolumn{2}{c|}{{\cellcolor[rgb]{0.937,0.937,0.937}}\en{Predicted}\de{Prognose }}           \\
\hhline{~~--|}
\multicolumn{1}{c}{}                                                             &                                                  & {\cellcolor[rgb]{0.937,0.937,0.937}}$\widehat{Y}$=1 & {\cellcolor[rgb]{0.937,0.937,0.937}}$\widehat{Y}$=0  \\ 
\hline
{\cellcolor[rgb]{0.937,0.937,0.937}}                                             & {\cellcolor[rgb]{0.937,0.937,0.937}}$Y$=1 & \en{True positives (TP)}\de{\makecell{Richtig-positive Prognosen\\ \textit{(True positives TP)}}}                              & \en{False negatives (FN)}\de{\makecell{Falsch-negative Prognosen\\ \textit{(False negatives FN)}}}                              \\ 
\hhline{|>{\arrayrulecolor[rgb]{0.937,0.937,0.937}}->{\arrayrulecolor{black}}---|}
\multirow{-2}{*}{{\cellcolor[rgb]{0.937,0.937,0.937}}\rotatebox[origin=c]{90}{\en{True}\de{Wahr}}} & {\cellcolor[rgb]{0.937,0.937,0.937}}$Y$=0 & \en{False positives (FP)}\de{\makecell{Falsch-positive Prognosen\\ \textit{(False positives FP)}}}                & \en{True negatives (TN)}\de{\makecell{Richtig-negative Prognosen\\ \textit{(True negatives TN)}}}                                \\
\hline
\end{tabular}
}

\newcommand{\confusionmatrixshort}[4]{%
\centering
\renewcommand\arraystretch{2}
\renewcommand\tabcolsep{10pt}
\arrayrulecolor{black}
\begin{tabular}{|c|c|c|c|} 
\hhline{~~--|}
\multicolumn{1}{c}{}                                                             &                                                  & \multicolumn{2}{c|}{{\cellcolor[rgb]{0.937,0.937,0.937}}\en{Predicted}\de{Prognose} }           \\
\hhline{~~--|}
\multicolumn{1}{c}{}                                                             &                                                  & {\cellcolor[rgb]{0.937,0.937,0.937}}$\widehat{Y}$=1 & {\cellcolor[rgb]{0.937,0.937,0.937}}$\widehat{Y}$=0  \\ 
\hline
{\cellcolor[rgb]{0.937,0.937,0.937}}                                             & {\cellcolor[rgb]{0.937,0.937,0.937}}$Y$=1 & #1                 & #2                            \\ 
\hhline{|>{\arrayrulecolor[rgb]{0.937,0.937,0.937}}->{\arrayrulecolor{black}}---|}
\multirow{-2}{*}{{\cellcolor[rgb]{0.937,0.937,0.937}}\rotatebox[origin=c]{90}{\en{True}\de{Wahr}}} & {\cellcolor[rgb]{0.937,0.937,0.937}}$Y$=0 & #3    & #4                            \\
\hline
\end{tabular}
}

%% file: macros/predictions.tex
\newcommand{\predictions}[4]{
\setlength{\tabcolsep}{1pt}

\resizebox{\textwidth}{!}{
\begin{tabular}{c|*{\the\numexpr#1+#2+#3+#4\relax}{c}}
$Y$ & 
\xintFor* ##9 in {\xintSeq {1}{\the\numexpr #3+#4 \relax}}\do {\xintifForFirst{}{&}0} & 
\xintFor* ##9 in {\xintSeq {1}{\the\numexpr #1+#2 \relax}}\do {\xintifForFirst{}{&}1}\\  
\hline
\rule{0pt}{12pt}$\widehat{Y}$ & 

\xintFor* ##9 in {\xintSeq {1}{\the\numexpr#4\relax}}\do {\xintifForFirst{}{&}0} & 

\xintFor* ##9 in {\xintSeq {1}{\the\numexpr#3\relax}}\do {\xintifForFirst{}{&}1} & 

\xintFor* ##9 in {\xintSeq {1}{\the\numexpr#1\relax}}\do {\xintifForFirst{}{&}1} &

\xintFor* ##9 in {\xintSeq {1}{\the\numexpr#2\relax}}\do {\xintifForFirst{}{&}0}

 \\ 

\multicolumn{1}{c}{} & 

\raisebox{5pt}{\tikzmark{tn1}} & \multicolumn{\the\numexpr #4-1 \relax}{r}{\raisebox{5pt}{\tikzmark{tn2}}} &

\raisebox{5pt}{\tikzmark{fp1}} &
\multicolumn{\the\numexpr #3-1 \relax}{r}{\raisebox{5pt}{\tikzmark{fp2}}} &

\raisebox{5pt}{\tikzmark{tp1}} &
\multicolumn{\the\numexpr #1-1 \relax}{r}{\raisebox{5pt}{\tikzmark{tp2}}} &

\raisebox{5pt}{\tikzmark{fn1}} &
\multicolumn{\the\numexpr #2-1 \relax}{r}{\raisebox{5pt}{\tikzmark{fn2}}}

\end{tabular}

\\
\begin{tikzpicture}[remember picture,overlay,baseline=50pt]
    \draw[thick,decoration={brace, amplitude=12 pt, aspect = 0.5},decorate] 
      (pic cs:tn2)
-- node[below=5mm]{\en{True negatives}\de{Richtig-negative}}
      (pic cs:tn1); 
\end{tikzpicture}
\\
\begin{tikzpicture}[remember picture,overlay,baseline=0pt]
    \draw[thick,decoration={brace, amplitude=12 pt, aspect = 0.5},decorate] 
      (pic cs:fp2)
-- node[below=5mm]{\en{False positives}\de{Falsch-positive}}
      (pic cs:fp1); 
\end{tikzpicture}
\\
\begin{tikzpicture}[remember picture,overlay,baseline=0pt]
    \draw[thick,decoration={brace, amplitude=12 pt, aspect = 0.5},decorate] 
      (pic cs:tp2)
-- node[below=5mm]{\en{True positives}\de{Richtig-positive}}
      (pic cs:tp1);
\end{tikzpicture}
\\
\begin{tikzpicture}[remember picture,overlay,baseline=20pt]
    \draw[thick,decoration={brace, amplitude=12 pt, aspect = 0.5},decorate] 
      (pic cs:fn2)
-- node[below=5mm]{\en{False negatives}\de{Falsch-negative}}
      (pic cs:fn1); 
\end{tikzpicture}
}}